\documentclass[10pt,twocolumn,letter]{IEEEtran}

\usepackage{amsmath,graphicx}
\usepackage{amsfonts}
\usepackage{amssymb}
\usepackage{epsfig}
\usepackage{mathrsfs}
\usepackage{graphicx}
\usepackage{algorithm}
\usepackage{caption}
\usepackage{subcaption}
\usepackage{epstopdf}
\usepackage{balance}
\usepackage{color}
\usepackage{cite}

\DeclareMathOperator{\diag}{\mathrm{diag} \, }

\newsavebox\myboxA
\newsavebox\myboxB
\newlength\mylenA

\newcommand{\oline}[1]{\mkern 1.5mu\overline{\mkern-1.5mu#1\mkern-1.5mu}\mkern 1.5mu}

\def\mA{\mbox{$\mathbf{A}$}}
\def\mB{\mbox{$\mathbf{B}$}}

\def\mC{\mbox{$\mathbf{C}$}}
\def\mD{\mbox{$\mathbf{D}$}}

\def\mH{\mbox{$\mathbf{H}$}}
\def\mI{\mbox{$\mathbf{I}$}}

\def\mP{\mbox{$\mathbf{P}$}}
\def\mQ{\mbox{$\mathbf{Q}$}}

\def\mU{\mbox{$\mathbf{U}$}}

\def\mX{\mbox{$\mathbf{X}$}}
\def\mY{\mbox{$\mathbf{Y}$}}

\def\mSigma{\mbox{$\mathbf{\Sigma} \kern .08em$}}
\def\mLambda{\mbox{$\mathbf{\Lambda} \kern .08em$}}

\newcommand{\E}{{\cal E}}
\newcommand{\G}{{\cal G}}

\newcommand{\V}{{\cal V}}

\def\S{\text{\mbox{${\cal S}$}}}
\def\F{\text{\mbox{${\cal F}$}}}

\def\V{\text{\mbox{${\cal V}$}}}
\def\E{\text{\mbox{${\cal E}$}}}
\def\G{\text{\mbox{${\cal G}$}}}

\def\bC{\text{\mbox{\boldmath $C$}}}

\def\bI{\text{\mbox{\boldmath $I$}}}

\def\b0{\text{\mbox{\boldmath $0$}}}

\def\bp{\text{\mbox{\boldmath $p$}}}

\def\bs{\text{\mbox{\boldmath $s$}}}
\def\bu{\text{\mbox{\boldmath $u$}}}

\def\bx{\text{\mbox{\boldmath $x$}}}
\def\by{\text{\mbox{\boldmath $y$}}}
\def\bz{\text{\mbox{\boldmath $z$}}}
\def\buno{\text{\mbox{\boldmath $1$}}}

\def\bv{\text{\mbox{\boldmath $v$}}}

\def\bpsi{\text{\mbox{\boldmath $\psi$}}}
\def\bPsi{\text{\mbox{\boldmath $\Psi$}}}
\newtheorem{theorem}{Theorem}
\newenvironment{proof}[1][Proof]{\noindent \textbf{#1.} }{\qedsymbol}
\newcommand{\qedsymbol}{\hspace{\fill}\rule{1.5ex}{1.5ex}}

\begin{document}

\title{Adaptive Graph Signal Processing:\\ Algorithms and Optimal Sampling Strategies}


\author{Paolo~Di Lorenzo,~\IEEEmembership{Member,~IEEE}, Paolo Banelli,~\IEEEmembership{Member,~IEEE}, Elvin Isufi,~\IEEEmembership{Student Member,~IEEE},\\ Sergio Barbarossa,~\IEEEmembership{Fellow,~IEEE}, and Geert Leus,~\IEEEmembership{Fellow,~IEEE}
\thanks{Di Lorenzo and Banelli are with the Dept. of Engineering, University of Perugia, Via G. Duranti 93, 06125, Perugia, Italy; Email: \texttt{paolo.dilorenzo@unipg.it, paolo.banelli@unipg.it}. \newline \indent Isufi and Leus are with EEMCS,  Delft  University of  Technology, 2826  CD  Delft,  The  Netherlands. \texttt{e.isufi-1@tudelft.nl,g.j.t.leus@tudelft.nl}.\newline \indent Barbarossa is with the Dept. of Information Engineering, Electronics, and Telecommunications, Sapienza University of Rome, Via Eudossiana 18, 00184, Rome, Italy; E-mail: \texttt{sergio.barbarossa@uniroma1.it}. The work of Paolo Di Lorenzo was supported by the ``Fondazione Cassa di Risparmio di Perugia''.}\vspace{-.6cm}
}

\maketitle

\begin{abstract}
The goal of this paper is to propose novel strategies for adaptive learning of signals defined over graphs, which are observed over a (randomly time-varying) subset of vertices. We recast two classical adaptive algorithms in the graph signal processing framework, namely, the least mean squares (LMS) and the recursive least squares (RLS) adaptive estimation strategies. For both methods, a detailed mean-square analysis illustrates the effect of random sampling on the adaptive reconstruction capability and the steady-state performance. Then, several probabilistic sampling strategies are proposed to design the sampling probability at each node in the graph, with the aim of optimizing the tradeoff between steady-state performance, graph sampling rate, and convergence rate of the adaptive algorithms. Finally, a distributed RLS strategy is derived and is shown to be convergent to its centralized counterpart. Numerical simulations carried out over both synthetic and real data illustrate the good performance of the proposed sampling and reconstruction strategies for (possibly distributed) adaptive learning of signals defined over graphs.
\end{abstract}


\section{Introduction}

In a large number of applications involving sensor, transportation, communication, social, or biological networks, the observed data can be modeled as signals defined over graphs, or graph signals for short. As a consequence, over the last few years, there was a surge of interest in developing novel analysis methods for graph signals, thus leading to the research field known as graph signal processing (GSP), see, e.g., \cite{shuman2013emerging,sandryhaila2013discrete,sandryhaila2014big}. The goal of GSP is to extend classical processing tools to the analysis of signals defined over an irregular discrete domain, represented by a graph, and one interesting aspect is that such methods typically come to depend on the graph topology, see, e.g., \cite{sandryhaila2013discrete}, \cite{sandryhaila2014discrete,narang2012perfect,narang2013compact,isufi2017autoregressive}. Probably, the most important processing tool is represented by graph spectral analysis, which hinges on the definition of the graph Fourier transform (GFT). The GFT was defined in two alternative ways, i.e., the projection of the graph signal onto the eigenvectors of either the Laplacian, see, e.g., \cite{shuman2013emerging}, \cite{pesenson2008sampling}, \cite{zhu2012approximating}, or of the adjacency matrix, see, e.g. \cite{sandryhaila2013discrete}, \cite{chen2015discrete}.
Quite recently, a third approach was also proposed, which builds the GFT as the set of orthonormal vectors that minimizes the Lov\'{a}sz extension of the graph cut size \cite{sardellitti2017graph}.

A fundamental task in GSP is to infer the attributes of some vertices from the observation of other vertices. In the GSP literature, this learning task is known as \textit{interpolation from samples}, and emerges whenever cost constraints limit the number of vertices that we can directly observe. An important contribution to sampling theory in GSP is given by \cite{pesenson2008sampling}, later extended in \cite{narang2013signal} and, very recently, in \cite{chen2015discrete,tsitsvero2015signals,wang2014local,marques2016sampling,gama2016rethinking}. Then, several interpolation methods were proposed to reconstruct signals on graphs from a subset of samples. Machine learning methods typically exploit smoothness of the signal over the graph, see, e.g., \cite{lafferty2002diffusion,chapelle2009semi,belkin2006manifold,romero2017kernel}, whereas GSP usually considers estimators for graph signals that adhere to the bandlimited model, i.e., signals that belong to the span of some eigenvectors of the graph Laplacian or  adjacency matrices, see, e.g., \cite{chen2015discrete},\cite{tsitsvero2015signals}, \cite{gama2016rethinking}. Sampling theory for graph signals will represent the basic tool to enable proactive sensing of large-scale cyber physical systems, with the aim of ensuring proper control of the system within its physical constraints and guaranteing a target level of performance, by only checking a limited number of nodes, over a limited number of time instants. In fact, as illustrated in several recent works as, e.g., \cite{chen2015discrete}, \cite{tsitsvero2015signals}, \cite{chamon2017greedy,chen2015signalrecovery,anis2016efficient}, when sampling graph signals, what matters is not only how many samples are taken but, most importantly, \textit{where} such samples are collected over the graph, thus catalyzing the interest for designing novel criteria and algorithms to select the set of sampled vertices.

In many applications such as, e.g., transportation networks, brain networks, or communication networks, the observed graph signals are typically time-varying. This requires the development of effective methods capable to learn and track dynamic graph signals from a carefully designed, possibly time-varying, sampling set. Some previous works have considered this specific learning task, see, e.g., \cite{dilo2016adaptive,di2016distAdaGraph,romero2016kernel,wang2015distributed}. Specifically, \cite{dilo2016adaptive} proposed an LMS estimation strategy enabling adaptive learning and tracking from a limited number of smartly sampled observations. The LMS method in  \cite{dilo2016adaptive} was then extended to the distributed setting in \cite{di2016distAdaGraph}. The work in \cite{romero2016kernel} proposed a kernel-based reconstruction framework to accommodate time-evolving signals over possibly time-evolving topologies, leveraging spatio-temporal dynamics of the observed data. Finally, reference \cite{wang2015distributed} proposes a distributed method for tracking
bandlimited graph signals, assuming perfect observations (i.e., there is no measurement noise) and a fixed sampling strategy.

The goal of this paper is to provide fundamental contributions toward the development of an adaptive graph signal processing framework, whose aim is to extend classical adaptive processing methods for the analysis of signals defined over graphs. The
main contributions of this work are summarized in the following two points.
\begin{enumerate}
  \item Similarly to the distributed case in \cite{di2016distAdaGraph}, we extend the LMS algorithm of \cite{dilo2016adaptive} to incorporate a \textit{probabilistic sampling} mechanism, where each node of the graph, at every time instant, is sampled with a given probability. Then, we derive a mean-square analysis that illustrates the role played by the sampling probabilities on the reconstruction capabilities and performance of the LMS algorithm. On the basis of the developed analysis, we design probabilistic sampling strategies aimed at minimizing the graph sampling rate (or maximizing the mean-square performance) while imposing accuracy (and/or sampling) constraints.
  \item We propose an RLS reconstruction strategy that collects data over the graph by the same probabilistic sampling method. Then, after giving necessary and sufficient conditions for adaptive graph signal reconstruction, we formulate an optimization problem to design the sampling probability at each node in the graph, reducing the sampling rate from one side, while also guaranteeing signal reconstruction and a prescribed steady-state performance. Finally, we derive a distributed RLS strategy for graph signals exploiting the alternating direction method of multipliers (ADMM) \cite{boyd2011distributed}.
\end{enumerate}

The proposed methods exploit the graph structure that describes the observed signal and, under a bandlimited assumption, enable adaptive reconstruction and tracking from a limited number of observations taken over a (possibly time-varying) subset of vertices. An interesting feature of our strategies is that this subset is allowed to vary over time, provided that the \textit{average} sampling set satisfies specific conditions enabling graph signal recovery.

The rest of the paper is organized as follows. In Section II, we summarize some GSP tools that will be used throughout the paper. Section III introduces the proposed LMS algorithm for adaptive learning of graph signals, illustrating the conditions enabling signal reconstruction from a (time-varying) subset of samples, deriving a detailed mean-square analysis, and developing useful sampling strategies. Similarly, Section IV considers RLS on graphs, along with its mean-square properties, sampling strategies, and distributed implementation. Then, in Section V we report several numerical simulations, aimed at assessing the validity of the theoretical analysis and the performance of the proposed algorithms. Finally, Section VI draws some conclusions.

\section{Background on Graph Signal Processing}

Let us consider a graph $\G = (\V, \E)$ consisting of a set of $N$ nodes $\V = \{1,2,..., N\}$, along with a set of weighted edges $\E=\{a_{ij}\}_{i, j \in \V}$, such that $a_{ij}>0$, if there is a link from node $j$ to node $i$, or $a_{ij}=0$, otherwise. The adjacency matrix $\mA$ of a graph is the collection of all the weights, i.e., $\mA=\{a_{ij}\}, i, j = 1, \ldots, N$.
The Laplacian matrix is defined as $\mathbf{L} = {\rm diag}(\mathbf{1}^T\mathbf{A})\!-\!\mathbf{A}$, where ${\rm diag}(\bx)$ is a matrix having $\bx$ as main diagonal, and zeros elsewhere. If the graph is {\it undirected}, the Laplacian matrix is symmetric and positive semi-definite, and can be decomposed as $\mathbf{L}=\mathbf{U}\boldsymbol{\Lambda}\mathbf{U}^H$, where $\mathbf{U}$ collects an orthonormal set of eigenvectors of $\mathbf{L}$ in its columns, whereas $\boldsymbol{\Lambda}$ is a diagonal matrix containing the real eigenvalues of $\mathbf{L}$.

A signal $\bx$ over a graph $\G$ is defined as a mapping from the vertex set to the set of complex numbers, i.e., $\bx: \V \rightarrow \mathbb{C}$.
The GFT $\bs$ of a signal $\bx$ is defined as the projection onto the set of vectors $\mU =\{\bu_i\}_{i=1,\ldots,N}$, i.e.,
\begin{equation}
\label{GFT}
\mathbf{\bs} = \mathbf{U}^H \bx,
\end{equation}
where $\{\bu_i\}_{i=1,\ldots,N}$ form an orthonormal basis and are commonly chosen as the eigenvectors
of either the graph Laplacian \cite{shuman2013emerging}, or of the adjacency matrix \cite{sandryhaila2013discrete}, \cite{chen2015discrete} (always assuming these matrices are normal). In this paper, we basically follow the approach from \cite{shuman2013emerging}, but the theory can be directly extended to other cases by simply substituting in (\ref{GFT}) the corresponding GFT basis.
We denote the support of $\bs$ in (\ref{GFT}) as
$\mathcal{F}=\{i\in\{1,\ldots,N\}:s_i\neq0\},$
and the \textit{bandwidth} of the graph signal $\bx$ is defined as the cardinality of $\mathcal{F}$, i.e., $|\mathcal{F}|$. The space of all signals whose GFT is supported on the set $\F$ is known as the {\it Paley-Wiener space} for the set $\F$ \cite{pesenson2008sampling}.

Finally, given a subset of vertices $\S \subseteq \V$, we define a vertex-limiting operator as the diagonal matrix
\begin{equation}
\label{D}
\mathbf{D}_{\S} = {\rm diag}\{\buno_{\S}\},
\end{equation}
where $\buno_{\S}$ is the set indicator vector, whose $i$-th entry is equal to one, if  $i \in \S$, or zero otherwise. Similarly, given a subset of frequency indices $\F\subseteq \V$, we introduce the filtering operator
\begin{equation}
\label{lowpass_operator}
\mB_{\F} = \mU_{\F}\mathbf{U}^H_{\F},
\end{equation}
where $\mU_{\F}\in \mathbb{C}^{N\times |\F|}$ collects the subset of columns of matrix $\mU$ in (\ref{GFT}) associated to the frequency indices $\F$. It is immediate to check that both matrices $\mathbf{D}_{\S}$ and $\mathbf{B}_{\F}$ are self-adjoint and idempotent, and so they represent orthogonal projectors onto the set of vertices $\S$ and the Paley-Wiener space $\F$, respectively.

\section{Adaptive Least Mean Squares Estimation\\ of Graph Signals}

Let us consider a signal $\bx^o=\{x^o_i\}_{i=1}^N\in\mathbb{C}^N$ defined over the graph $\G = (\V, \E)$. To enable sampling of $\bx^o$ without loss of information, the following is assumed:

\vspace{.1cm}
\textit{Assumption 1 (Bandlimited): The signal $\bx^o$ is $\F$-bandlimited over the graph $\mathcal{G}$, i.e., its spectral content is different from zero only on the set of indices $\F$.} \qedsymbol
\vspace{.1cm}

If the signal support $\F$ is fixed and known beforehand, from (\ref{GFT}), the graph signal $\bx^o$ can be modeled in compact form as:
\begin{equation}
\label{compact_decomp}
\bx^o=\mU_{\F}\bs^o,
\end{equation}
where $\bs^o\in\mathbb{C}^{|\F|}$ is the vector of GFT coefficients of the frequency support of the graph signal $\bx^o$.  At each time $n$, noisy samples of the signal are taken over a (randomly) time-varying subset of vertices, according to the following model:
\begin{align}
\label{lin_observation}
\by[n]\,=\,&\mD_{\S[n]}\left(\bx^o+\bv[n]\right)=\,\mD_{\S[n]}\mU_{\F}\bs^o+\mD_{\S[n]}\bv[n]
\end{align}
where $\mD_{\S[n]}={\rm diag}\{d_1[n],\ldots,d_N[n]\}\in\mathbb{R}^{N\times N}$ [cf. (\ref{D})], with $d_i[n]$ denoting a random sampling binary coefficient, which is equal to 1 if $i\in \S[n]$, and 0 otherwise (i.e., $\S[n]$ represents the \textit{instantaneous}, random sampling set at time $n$); and $\bv[n]\in\mathbb{C}^N$ is zero-mean, spatially and temporally independent observation noise, with covariance matrix $\mC_v=\diag\{\sigma_1^2,\ldots,\sigma_N^2\}$. The estimation task consists in recovering the vector $\bx^o$ (or, equivalently, its GFT $\bs^o$) from the noisy, streaming, and partial observations $\by[n]$ in (\ref{lin_observation}). Following an LMS approach \cite{sayed2011adaptive}, from (\ref{lin_observation}), the optimal estimate for $\bs^o$ can be found as the vector that solves the optimization problem\footnote{Note that, since $\mU_{\F}$ has orthonormal columns, estimating $\bs^o$ or $\bx^o$ is equivalent from a mean square error perspective.}:
\begin{align}
\label{LMS_problem}
&\min_{\boldsymbol{s}} \; \mathbb{E}\, \|\mD_{\S[n]}(\by[n]-\mU_{\F}\bs)\|^2
\end{align}
where $\mathbb{E}(\cdot)$ denotes the expectation operator, and in (\ref{LMS_problem}) we have exploited the fact that $\mD_{\S[n]}$ is an idempotent matrix for any fixed $n$ [cf. (\ref{D})]. A typical LMS-type solution proceeds to optimize (\ref{LMS_problem}) by means of a stochastic steepest-descent procedure, relying only on instantaneous information. Thus, letting $\widehat{\bs}[n]$ be the current estimate of vector $\bs^o$, the LMS algorithm for graph signals evolves as:
\begin{align}\label{LMS_GFT}
\widehat{\bs}[n+1]=\widehat{\bs}[n]+\mu\,\mU_{\F}^H\mD_{\S[n]}\left(\by[n]-\mU_{\F}\widehat{\bs}[n]\right),
\end{align}
where $\mu>0$ is a (sufficiently small) step-size, and we have exploited the fact that $\mD_{\S[n]}$ is an idempotent operator. Finally, exploiting (\ref{compact_decomp}) and (\ref{lowpass_operator}), recursion (\ref{LMS_GFT}) can be equivalently recast with respect to the current estimates of graph signal $\bx^o$, say $\widehat{\bx}[n]$, as illustrated in Algorithm 1.
At every iteration $n$, Algorithm 1 evolves implementing an alternating orthogonal projection onto the instantaneous sampling set $\S[n]$ (through $\mD_{\S[n]}$) and the Paley-Wiener space $\F$ (through $\mB_{\F}$), with an overall complexity given by $O(|\F||\S[n]|)$. The LMS strategy in Algorithm 1 extends the method previously proposed in \cite{dilo2016adaptive} by incorporating the random sampling mechanism defined by the (time-varying) sampling operator $\mD_{\S[n]}$. Of course, the mean-square properties of the LMS recursion crucially depend on the properties of the random sampling operator $\mD_{\S[n]}$. Thus, in the sequel, we will show how the design of the sampling strategy affects the reconstruction capability, the learning rate, and the steady-state performance of Algorithm 1. Before moving forward, we introduce an independence assumption on the random sampling process.

{\textit{Assumption 2 (Independent sampling)}}:  The random variables extracted from the sampling process $\{d_i[l]\}$ are temporally and spatially independent, for all $i$, and $l\leq n$. \qedsymbol

\subsection{Adaptive Reconstruction from Sparse Sampling} \label{AdaptRecSparseSampling}

Assuming the random sampling and observation processes $\{d_i[n]\}_{i=1}^N$ and $\{\by[n]\}$ to be stationary over time, the solution of problem (\ref{LMS_problem}) is given by the vector $\bs^o$ that satisfies:
\begin{align}
\label{normal_equations}
\mU_{\F}^H\,{\rm diag}(\bp)\mU_{\F} \bs^o = \mU_{\F}^H\,\mathbb{E}\{\by[n]\},
\end{align}
where $\bp=(p_1,\ldots,p_N)^T\in\mathbb{R}^{N}$ represents the sampling probability vector, with $p_i=\mathbb{E}\{d_i[n]\}$, $i=1,\ldots,N$, denoting the probability that node $i$ is sampled at time $n$. The system of equations (\ref{normal_equations}) admits a unique solution $\bs^o$ if the square matrix $\mU_{\F}^H\,{\rm diag}(\bp)\mU_{\F}$ is full rank or, equivalently, if
\begin{equation}\label{lambda_min}
   \lambda_{\min}\left(\mU_{\F}^H\,{\rm diag}(\bp)\mU_{\F}\right)>0,
\end{equation}
where $\lambda_{\min}(\mY)$ is the minimum eigenvalue of matrix $\mY$. Also, let us denote the \textit{expected sampling set} by $$\overline{\S}=\{i=1,\ldots,N \,|\, p_i>0\},$$ i.e., the set of nodes of the graph that are sampled with a probability strictly greater than zero.
\begin{algorithm}[t]
\caption*{\textbf{Algorithm 1: LMS on Graphs}}
\vspace{.1cm}
Start with random $\widehat{\bx}[0]$. Given a sufficiently small step-size $\mu>0$, for each time $n\geq0$, repeat:
\begin{equation}\label{LMS}
\widehat{\bx}[n+1]=\widehat{\bx}[n]+\mu\,\mB_{\F}\mD_{\S[n]}\left(\by[n]-\widehat{\bx}[n]\right)\nonumber\\
\end{equation}
\end{algorithm}
To provide a necessary and sufficient condition for signal reconstruction, we proceed similarly to \cite{tsitsvero2015signals,di2016distAdaGraph}. Since $p_i>0$ for all $i\in \overline{\mathcal{S}}$,
\begin{align}\label{cond2}
\mathrm{rank}\left(\mU_{\F}^H\,{\rm diag}(\bp)\mU_{\F}\right)=\mathrm{rank}\left(\mU_{\F}^H\mD_{\,\overline{\mathcal{S}}}\mU_{\F}\right),
\end{align}
i.e., condition (\ref{lambda_min}) holds true if matrix
$\mU_{\F}^H\mD_{\,\overline{\mathcal{S}}}\mU_{\F}$
has full rank, where $\mD_{\,\overline{\mathcal{S}}}$ is a vertex-limiting operator that projects onto the expected sampling set $\overline{\S}$ [cf. (\ref{D})]. Let us now introduce
$\mD_{\,\overline{\mathcal{S}}_c}=\mI-\mD_{\,\overline{\mathcal{S}}}$,
which projects onto the complement of the expected sampling set, i.e. $\overline{\S}_c=\mathcal{V} \setminus\overline{\mathcal{S}}=\{i=1,\ldots,N \,|\, p_i=0\}$. Then, signal reconstruction is possible if
$\mU_{\F}^H\mD_{\,\overline{\mathcal{S}}}\mU_{\F}=\mI-\mU_{\F}^H\mD_{\,\overline{\mathcal{S}}_c}\mU_{\F}$
is invertible, i.e., if $\left\|\mU_{\F}^H\mD_{\,\overline{\mathcal{S}}_c}\mU_{\F}\right\|_2<1$ or, equivalently, if
\begin{equation}\label{|DcB|<1}
   \left\| \mD_{\,\overline{\mathcal{S}}_c}\mU_{\F}\right\|_2 < 1.
\end{equation}
As illustrated in \cite{tsitsvero2015signals,dilo2016adaptive}, condition (\ref{|DcB|<1}) implies that there are no $\F$-bandlimited signals that are perfectly localized over the set $\oline{\S}_c$, thus providing a \textit{necessary} and \textit{sufficient} condition for adaptive graph signal reconstruction. Furthermore, differently from previous works on sampling of graph signals, see, e.g.,
\cite{pesenson2008sampling,narang2013signal,chen2015discrete,tsitsvero2015signals,wang2014local,marques2016sampling,dilo2016adaptive}, condition (\ref{|DcB|<1}) depends on the \textit{expected} sampling set. As a consequence, the proposed LMS algorithm with probabilistic sampling does not need to collect all the data necessary to reconstruct one-shot the graph signal at each iteration, but can learn acquiring the needed information over time. The only important thing required by condition (\ref{|DcB|<1}) is that a sufficiently large number of nodes is sampled in \textit{expectation} (i.e., the expected sampling set $\overline{\mathcal{S}}$ contains a sufficiently large number of nodes).

\subsection{Mean-Square Performance}
\label{LMS_analysis}

We now illustrate how the sampling probability vector $\bp$ affects the mean-square behavior of Algorithm 1. Let $\widetilde{\bx}[n]=\widehat{\bx}[n]-\bx^o$ and $\widetilde{\bs}[n]=\widehat{\bs}[n]-\bs^o$ be the error vectors on the graph signal and its GFT, respectively, at time $n$. Thus, using (\ref{lin_observation}) in (\ref{LMS_GFT}), we obtain:
\begin{equation}
\label{error_recursion}
\widetilde{\bs}[n+1]=(\mI-\mu\,\mU_{\F}^H\mD_{\S[n]}\mU_{\F})\,\widetilde{\bs}[n]+\mu\,\mU_{\F}^H\mD_{\S[n]} \bv[n].
\end{equation}
Starting from (\ref{error_recursion}), it is possible to derive a complete mean-square analysis of Algorithm 1, which relies also on the following assumption.

{\textit{Assumption 3 (Small step-size)}}: {The step-size $\mu$ is chosen sufficiently small so that terms that depend on higher-order powers of $\mu$ can be ignored.}{\qedsymbol}

\noindent The main results are summarized in the following Theorem.

\textit{\begin{theorem}\label{theorem_mean_square}
Given model (\ref{lin_observation}), under Assumptions 2 and 3, and for any initial condition, Algorithm 1 is stable in the mean-square error sense if the sampling probability vector $\bp$ and the step-size $\mu$ are chosen to satisfy (\ref{|DcB|<1}) and
\begin{equation}\label{step}
0< \mu < \frac{2\lambda_{\min}\left(\mU_{\F}^H\,{\rm diag}(\bp)\mU_{\F}\right)}{\lambda^2_{\max}\left(\mU_{\F}^H\,{\rm diag}(\bp)\mU_{\F}\right)}.
\end{equation}
Furthermore, the mean square deviation (MSD) writes as \footnote{Note that, if (\ref{|DcB|<1}) holds and considering spatially white noise samples with equal variance $\sigma_v^2$, expression \eqref{MSD} reduces to ${\rm MSD}=\frac{\mu}{2}|\F| \sigma_v^2+O(\mu^2)$ .}
\begin{align}\label{MSD}
&{\rm MSD}=\lim_{n\rightarrow\infty}\,\mathbb{E}\|\widetilde{\bx}[n]\|^2 \,=\, \lim_{n\rightarrow\infty}\,\mathbb{E}\|\widetilde{\bs}[n]\|^2 \\
&=\,\frac{\mu}{2}\,{\rm Tr}\left[\left(\mU_{\F}^H\,{\rm diag}(\bp)\mU_{\F}\right)^{-1}\mU_{\F}^H\,{\rm diag}(\bp)\mC_v\mU_{\F}\right] +O(\mu^2)  \nonumber
\end{align}
and the convergence rate $\alpha$ is well approximated by
\begin{equation}\label{learning_rate}
\alpha\simeq
 1-2\mu \lambda_{\min}\left(\mU_{\F}^H\,{\rm diag}(\bp)\mU_{\F}\right)
\end{equation}
when $\displaystyle\mu \ll \frac{2\lambda_{\rm{min}}\left(\mU_{\F}^H\,{\rm diag}(\bp)\mU_{\F}\right)}{\lambda_{\rm{max}}^2\left(\mU_{\F}^H\,{\rm diag}(\bp)\mU_{\F}\right)}.$
\end{theorem}} \smallskip

\begin{proof}
See Appendix A.
\end{proof} \smallskip


\subsection{Optimal Sampling Strategies}

The mean-square analysis in Sec. III-B illustrates how the convergence rate and the mean-square performance of Algorithm 1 depend on the sampling probability vector $\bp$ [cf. (\ref{MSD}) and (\ref{learning_rate})]. Then, following a sparse sensing approach \cite{chepuri2015sparsity,chepuri2016sparse}, the goal of this section is to develop optimal sampling strategies aimed at designing the probability vector $\bp$ that optimizes the tradeoff between steady-state performance, graph sampling rate, and convergence rate of Algorithm 1. In the sequel, under Assumption 3, we neglect the term $O(\mu^2)$ in (\ref{MSD}), and consider (\ref{learning_rate}) as the convergence rate.

\subsubsection{Minimum Graph Sampling Rate subject to Learning Constraints}

The first sampling strategy aims at designing the probability vector $\bp$ that minimizes the total sampling rate over the graph, i.e., $\mathbf{1}^T\bp$, while guaranteeing a target performance in terms of MSD in (\ref{MSD}) and of convergence rate in (\ref{learning_rate}). The optimization problem can be cast as:
\begin{equation}\label{min_sampling_rate_problem}
\setlength{\jot}{6pt}
\begin{aligned}
&\hspace{-.3cm}\min_{\boldsymbol{p}} \;\; \mathbf{1}^T\bp  \\
&\hspace{-.1cm}\hbox{subject to} \\
&\;\lambda_{\min}\left(\mU_{\F}^H\,{\rm diag}(\bp)\mU_{\F}\right)\geq \displaystyle\frac{1-\bar{\alpha}}{2\mu} \\
&\;{\rm Tr}\left[\left(\mU_{\F}^H\,{\rm diag}(\bp)\mU_{\F}\right)^{-1}\mU_{\F}^H\,{\rm diag}(\bp)\mC_v\mU_{\F}\right]\leq\frac{2\gamma}{\mu}\\
&\;\mathbf{0}\leq \bp \leq \bp^{\max}
\end{aligned}
\end{equation}
The first constraint imposes that the convergence rate of the algorithm is larger than a desired value, i.e., $\alpha$ in (\ref{learning_rate}) is smaller than a target value, say, e.g., $\bar{\alpha}\in(0,1)$. Note that the first constraint on the convergence rate also guarantees adaptive signal reconstruction [cf. (\ref{lambda_min})]. The second constraint guarantees a target mean-square performance, i.e., the MSD in (\ref{MSD}) must be less than or equal to a prescribed value, say, e.g., $\gamma>0$. Finally, the last constraint limits the probability vector to lie in the box $p_i\in[0,p^{\max}_i]$, for all $i$, with $0\leq p^{\max}_i\leq 1$ denoting an upper bound on the sampling probability at each node that might depend on external factors such as, e.g., limited energy, processing, and/or communication resources, node or communication failures, etc.

Unfortunately, problem (\ref{min_sampling_rate_problem}) is non-convex, due to the presence of the non-convex constraint on the MSD. To handle the non-convexity of (\ref{min_sampling_rate_problem}), in the sequel we follow two different approaches.
In the first place, under Assumption 3 [i.e., neglecting the terms $O(\mu^2)$], we exploit an upper bound of the MSD function in (\ref{MSD}), given by:
\begin{equation}\label{MSD_bound}
{\rm MSD}(\bp)\,\leq\,\overline{{\rm MSD}}(\bp)\triangleq \frac{\mu}{2}\,\frac{{\rm Tr}\left(\mU_{\F}^H\,{\rm diag}(\bp)\mC_v\mU_{\F}\right)}{\lambda_{\min}\left(\mU_{\F}^H\,{\rm diag}(\bp)\mU_{\F}\right)},
\end{equation}
for all $\bp\in \mathbb{R}^N$. Of course, replacing the MSD function (\ref{MSD}) with the bound (\ref{MSD_bound}), the second constraint of problem (\ref{min_sampling_rate_problem}) is always satisfied. Thus, exploiting the upper bound (\ref{MSD_bound}), we formulate a surrogate optimization problem for the selection of the probability vector $\bp$, which can be cast as:
\begin{equation}\label{min_sampling_rate_problem2}
\setlength{\jot}{6pt}
\begin{aligned}
&\hspace{-.8cm}\min_{\boldsymbol{p}} \;\; \mathbf{1}^T\bp  \\
&\hspace{-.4cm}\hbox{subject to} \\
&\;\lambda_{\min}\left(\mU_{\F}^H\,{\rm diag}(\bp)\mU_{\F}\right)\geq \displaystyle\frac{1-\bar{\alpha}}{2\mu} \\
&\;\frac{{\rm Tr}\left(\mU_{\F}^H\,{\rm diag}(\bp)\mC_v\mU_{\F}\right)}{\lambda_{\min}\left(\mU_{\F}^H\,{\rm diag}(\bp)\mU_{\F}\right)}\leq\frac{2\gamma}{\mu}\\
&\;\mathbf{0}\leq \bp \leq \bp^{\max}
\end{aligned}
\end{equation}
Problem (\ref{min_sampling_rate_problem2}) is now a convex optimization problem. Indeed, the second constraint of problem (\ref{min_sampling_rate_problem2}) involves the ratio of a convex function over a concave function. Since both functions at numerator and denominator of (\ref{MSD_bound}) are differentiable and positive for all $\bp$ satisfying the first and third constraint of  problem (\ref{min_sampling_rate_problem2}), the function is pseudo-convex \cite{avriel2010generalized}, and all its sub-level sets are convex sets. This argument, coupled with the convexity of the objective function and of the sets defined by the first and third constraints, proves the convexity of the problem in (\ref{min_sampling_rate_problem2}), whose global solution can be found using efficient numerical tools \cite{boyd2004convex}.

The second approach exploits successive convex approximation (SCA) methods \cite{scutari2017parallel}, whose aim is to find local optimal solutions of (\ref{min_sampling_rate_problem}). The issue in (\ref{min_sampling_rate_problem}) is the non-convexity of the feasible set. Thus, following the approach from \cite{scutari2017parallel}, we define the (non-convex) set $\mathcal{T}$ and the (convex) set $\mathcal{K}$ as:
\begin{align}\label{set_kappa}
\setlength{\jot}{6pt}
\mathcal{T}\triangleq
\begin{cases}
\lambda_{\min}\left(\mU_{\F}^H\,{\rm diag}(\bp)\mU_{\F}\right)\geq \displaystyle\frac{1-\bar{\alpha}}{2\mu}   \nonumber\\
\displaystyle\frac{\mu}{2}\displaystyle{\rm Tr}\left[\left(\mU_{\F}^H\,{\rm diag}(\bp)\mU_{\F}\right)^{-1}\mU_{\F}^H\,{\rm diag}(\bp)\mC_v\mU_{\F}\right]\leq \gamma\nonumber\\
\mathbf{0}\leq \bp \leq \bp^{\max}\nonumber
\end{cases}
\end{align}
and
\begin{equation}\label{set_C}
\setlength{\jot}{6pt}
\mathcal{K}\triangleq
\begin{cases}
\lambda_{\min}\left(\mU_{\F}^H\,{\rm diag}(\bp)\mU_{\F}\right)\geq \displaystyle\frac{1-\bar{\alpha}}{2\mu}    \nonumber\\
\mathbf{0}\leq \bp \leq \bp^{\max}\nonumber
\end{cases}
\end{equation}
Then, we replace the (second) non-convex constraint in $\mathcal{T}$ with a convex approximation
$\widetilde{\rm MSD}(\bp; \bz): \mathcal{K}\times\mathcal{T}\rightarrow \mathbb{R}$ that satisfies the following conditions:
 \begin{description}
\item[\hspace{-.2cm}(C1)] \hspace{-.2cm}
 $\widetilde{\rm MSD}(\mathbf{\bullet}; \bz)$ is convex on $\mathcal K$ for all $\bz\in \mathcal T$;\smallskip
\item[\hspace{-.2cm}(C2)] \hspace{-.2cm} $\widetilde{\rm MSD}(\bz;\bz) = {\rm MSD}(\bz)$ for all $\bz \in \mathcal T$;\smallskip
\item[\hspace{-.2cm}(C3)] \hspace{-.2cm} ${\rm MSD}(\bp)\leq\widetilde{\rm MSD}(\bp;\bz)$ for all $\bp\in \mathcal K$ and $\bz\in \mathcal T$; \smallskip
\item[\hspace{-.2cm}(C4)] \hspace{-.2cm} $\widetilde{\rm MSD}(\mathbf{\bullet}; \mathbf{\bullet})$ is continuous on ${\cal K} \times {\cal T}$;
\item[\hspace{-.2cm}(C5)] \hspace{-.2cm} $\nabla_{\bp}{\rm MSD}(\bz)=\nabla_{\bp}\widetilde{\rm MSD}(\bz;\bz)$ for all $\bz \in \mathcal T$;
\item[\hspace{-.2cm}(C6)] \hspace{-.2cm} $\nabla_{\bp}\widetilde{\rm MSD}(\mathbf{\bullet}; \mathbf{\bullet})$ is continuous on ${\cal K} \times {\cal T}$;
\end{description}
where $\nabla_{\bp}\widetilde{\rm MSD}(\bz;\bz)$ denotes the partial gradient of $\widetilde{\rm MSD}$ with respect to the first argument evaluated at $\bz$ (the second argument is kept fixed at $\bz$). Among the possible choices for $\widetilde{\rm MSD}$ (see \cite{scutari2017parallel} for details), since the ${\rm MSD}$ function in (\ref{MSD}) has Lipshitz continuous gradient on $\mathcal K$ (with Lipshitz constant $L$), a possible choice for the approximation is given by:
\begin{equation}\label{approx2}
\widetilde{\rm MSD}(\bp;\bz) = {\rm MSD}(\bz)+\nabla_{\bp}{\rm MSD}(\bz)^T(\bp-\bz)+\frac{L}{2}\|\bp-\bz\|^2,
\end{equation}
with $\bp \in \mathcal K$, $\bz \in \mathcal T$, and satisfying all conditions C1-C6. Thus, given the current estimate of $\bp$ at time $k$, say $\bp[k]$, we define a surrogate optimization set $\widetilde{{\cal T}}(\bp[k])$ given by:
\begin{align}\label{set_kappa_surrogate}
\setlength{\jot}{6pt}
\widetilde{{\cal T}}(\bp[k])\triangleq
\begin{cases}
\lambda_{\min}\left(\mU_{\F}^H\,{\rm diag}(\bp)\mU_{\F}\right)\geq \displaystyle\frac{1-\bar{\alpha}}{ 2\mu}   \nonumber\\
\widetilde{\rm MSD}(\bp;\bp[k])\leq \gamma\nonumber\\
\mathbf{0}\leq \bp \leq \bp^{\max}\nonumber
\end{cases}
\end{align}
with $\widetilde{\rm MSD}(\bp;\bp[k])$ given by (\ref{approx2}). Then, the SCA algorithm for problem (\ref{min_sampling_rate_problem}) proceeds as described in Algorithm 2. The updating scheme reads: at every iteration $k$, given the current estimate $\bp[k]$, the first step of Algorithm 2 solves a surrogate optimization problem involving the objective function $\mathbf{1}^T\bp$, augmented with a proximal regularization term (with $\tau>0$), and the surrogate set $\widetilde{{\cal T}}(\bp[k])$. Then, the second step of Algorithm 2 generates the new point $\bp[k+1]$ as a convex combination of the current estimate $\bp[k]$ and the solutions $\widehat{\bp}[k]$ of the surrogate problem. Under mild conditions on the step-size sequence $\gamma[k]$, and assuming $\widetilde{\mathcal{T}}(\bp[k])$ as a surrogate feasible set at each iteration $k$, the sequence generated by Algorithm 2 converges to a local optimal solution of problem (\ref{min_sampling_rate_problem}), see \cite[Theorem 2]{scutari2017parallel} for details.
\begin{algorithm}[t]
\caption*{\textbf{Algorithm 2: SCA method for Problem (\ref{min_sampling_rate_problem})}}
\vspace{.1cm}
Set $k=1$. Start with $\bp[1]\in \mathcal{T}$. Then, for $k\geq1$, repeat until convergence the following steps:
\begin{equation}
\setlength{\jot}{6pt}
\begin{aligned}
&{\rm S.1)}\;\;\;\widehat{\bp}[k]\;=\;\underset{\boldsymbol{p}\in \widetilde{{\cal T}}(\bp[k])}{\arg\min}\;\; \mathbf{1}^T\bp+\frac{\tau}{2}\|\bp-\bp[k]\|^2\nonumber\\
&{\rm S.2)}\;\;\;\bp[k+1]\;=\;\bp[k]+\gamma[k]\Big(\widehat{\bp}[k]-\bp[k]\Big)\nonumber
\end{aligned}
\end{equation}
\end{algorithm}

\subsubsection{Minimum MSD with Sampling and Learning Constraints}

The second sampling strategy aims at designing the probability vector $\bp$ that minimizes the MSD in (\ref{MSD}), while imposing that the convergence rate is larger than a desired value, and the total sampling rate $\mathbf{1}^T\bp$ is limited by a budget constraint. The optimization problem can then be cast as:

\begin{equation}\label{min_MSD_problem}
\setlength{\jot}{6pt}
\begin{aligned}
&\min_{\boldsymbol{p}} \;\; {\rm Tr}\left[\left(\mU_{\F}^H\,{\rm diag}(\bp)\mU_{\F}\right)^{-1}\mU_{\F}^H\,{\rm diag}(\bp)\mC_v\mU_{\F}\right]   \\
&\;\;\;\hbox{s.t.} \;\;
\bp\in\mathcal{C}\triangleq\begin{cases}
\lambda_{\min}\left(\mU_{\F}^H\,{\rm diag}(\bp)\mU_{\F}\right)\geq \displaystyle\frac{1-\bar{\alpha}}{2\mu} \\
\mathbf{1}^T\bp\leq \mathcal{P}\\
\mathbf{0}\leq \bp \leq \bp^{\max}
\end{cases}
\end{aligned}
\end{equation}
where $\mathcal{P}\in[0,\mathbf{1}^T\bp^{\max}]$ is the budget on the total sampling rate. Problem (\ref{min_MSD_problem}) has a convex feasible set $\mathcal{C}$, but it is non-convex because of the MSD objective function. Again, to handle the non-convexity of (\ref{min_MSD_problem}), we can follow two different paths.  Exploiting the bound in (\ref{MSD_bound}), it is possible to formulate an approximated optimization problem for the design of the probability vector $\bp$, which can be cast as:
\begin{equation}
\label{min_MSD_problem2}
\min_{\boldsymbol{p}\in\mathcal{C}} \;\; \frac{{\rm Tr}\left(\mU_{\F}^H\,{\rm diag}(\bp)\mC_v\mU_{\F}\right)}{\lambda_{\min}\left(\mU_{\F}^H\,{\rm diag}(\bp)\mU_{\F}\right)}.
\end{equation}
Problem (\ref{min_MSD_problem2}) is a convex/concave fractional program \cite{schaible2004recent}, i.e., a problem that involves the minimization of the ratio of a convex function over a concave function, both defined over the convex set $\mathcal{C}$. As previously mentioned, since both functions at numerator and denominator of (\ref{MSD_bound}) are differentiable and positive for all $\bp\in \mathcal{C}$, the objective function of (\ref{min_MSD_problem2}) is pseudo-convex in $\mathcal{C}$ \cite{avriel2010generalized}. As a consequence, any local minimum of problem (\ref{min_MSD_problem2}) is also a global minimum, and any descent method can be used to find such optimal solutions \cite{schaible2004recent}.
\begin{algorithm}[t]
\caption*{\textbf{Algorithm 3: Dinkelbach method for Problem (\ref{min_MSD_problem2})}}
\vspace{.1cm}
Set $k=1$. Start with $\bp[1]\in \mathcal{C}$ and $\omega[1]=\kappa(\bp[1])$. Then, for $k\geq 1$, repeat the following steps:
\begin{equation}
\setlength{\jot}{6pt}
\begin{aligned}\label{Dinkelbach}
&\;\;{\rm S.1)}\;\;\; \bp[k+1]=\underset{\boldsymbol{p}\in \mathcal{C}}\arg\min\; h(\bp,\omega[k]) \\
&\;\;{\rm S.2)}\;\;\; \hbox{If $h(\bp[k+1],\omega[k])=0$, STOP and $\bp^*=\bp[k+1]$;}\\
&\hbox{otherwise, $\omega[k+1]=\kappa(\bp[k+1])$, $k=k+1$, and go to S.1} \nonumber
\end{aligned}
\end{equation}
\end{algorithm}

To solve problem (\ref{min_MSD_problem2}), in this paper we consider a method based on the Dinkelbach algorithm \cite{dinkelbach1967nonlinear}, which converts the fractional problem (\ref{min_MSD_problem2}) into the iterative solution of a sequence of parametric problems as:
\begin{align}\label{parametric}
\min_{\boldsymbol{p}\in \mathcal{C}}\; h(\bp,\omega)\;=\; f(\bp)-\omega g(\bp)
\end{align}
with $\omega$ denoting the free parameter to be selected, and
\begin{align}
&f(\bp)={\rm Tr}\left(\mU_{\F}^H\,{\rm diag}(\bp)\mC_v\mU_{\F}\right),\\
&g(\bp)=\lambda_{\min}\left(\mU_{\F}^H\,{\rm diag}(\bp)\mU_{\F}\right).
\end{align}
Letting $\kappa(\bp)=f(\bp)/g(\bp)$, and noting that $h(\bp^*,\kappa(\bp^*))=0$ at the optimal value $\bp^*$, the Dinkelbach method proceeds as described in Algorithm 3, and is guaranteed to converge to global optimal solutions of the approximated problem (\ref{min_MSD_problem2}), see, e.g., \cite{dinkelbach1967nonlinear,schaible2004recent}.

The second approach aims at finding local optimal solutions of (\ref{min_MSD_problem}) using an SCA method with provable convergence guarantees. Following the approach proposed in \cite{scutari2017parallel}, and letting $\bp[k]$ be the guess of the probability vector at iteration $k$, the SCA algorithm proceeds as described in Algorithm 4. More formally, the updating scheme reads: at every iteration $k$, given the current estimate $\bp[k]$, the first step of Algorithm 4 solves a surrogate optimization problem involving the function $\widetilde{\rm MSD}(\bp;\bp[k])$, which represents a strongly convex approximant of ${\rm MSD}$ in (\ref{MSD}) at the point $\bp[k]$ that satisfies the following conditions:
 \begin{description}
\item[ \hspace{-.25cm}(F1)] \hspace{-.4cm}
 $\widetilde{\rm MSD}(\mathbf{\bullet}; \bz)$ is uniformly strongly convex on $\mathcal C$ $\forall\bz\in \mathcal{C}$;\smallskip
\item[ \hspace{-.4cm} (F2)] \hspace{-.4cm} $\nabla_{\bp} \widetilde{\rm MSD}(\bz;\bz) = \nabla_{\bp} {\rm MSD}(\bz)$ for all $\bz \in \mathcal C$;\smallskip
\item[ \hspace{-.4cm} (F3)] \hspace{-.4cm} $\nabla_{\bp} \widetilde{\rm MSD}(\mathbf{\bullet};\mathbf{\bullet})$ is continuous on $\mathcal C$.
\end{description}
Among all the possible choices satisfying the requirements F1-F3 (see \cite{scutari2017parallel} for details), in this paper we consider the following surrogate function:
\begin{equation}
\begin{aligned}
\label{surrogate}
&\hspace{-.2cm} \widetilde{\rm MSD}(\bp;\bp[k])= \frac{\tau}{2}\|\bp-\bp[k]\|^2\\
&\hspace{-.2cm}+\frac{\mu}{2} {\rm Tr}\left[\left(\mU_{\F}^H\,{\rm diag}(\bp[k])\mU_{\F}\right)^{-1}\mU_{\F}^H\,{\rm diag}(\bp)\mC_v\mU_{\F}\right]\\
&\hspace{-.2cm} +\frac{\mu}{2}{\rm Tr}\left[\left(\mU_{\F}^H\,{\rm diag}(\bp)\mU_{\F}\right)^{-1}\mU_{\F}^H\,{\rm diag}(\bp[k])\mC_v\mU_{\F}\right]
\end{aligned}
\end{equation}
with $\tau>0$, which satisfies F1-F3 and preserves much of the convexity hidden in the original function (\ref{MSD}). Then, the second step of Algorithm 4 generates the new point $\bp[k+1]$ as a convex combination of the current estimate $\bp[k]$ and the solutions $\widehat{\bp}[k]$ of the surrogate problem, exploiting the step-size sequence $\gamma[k]$. Under mild conditions on the step-size sequence $\gamma[k]$, and using (\ref{surrogate}), the sequence generated by Algorithm 4 converges to a local optimal solution of (\ref{min_MSD_problem}), see \cite[Theorem 2]{scutari2017parallel} for details.

\begin{algorithm}[t]
\caption*{\textbf{Algorithm 4: SCA method for Problem (\ref{min_MSD_problem})}}
\vspace{.1cm}
Set $k=1$. Start with $\bp[1]\in \mathcal{C}$. Then, for $k\geq1$, repeat until convergence the following steps:
\begin{equation}
\setlength{\jot}{6pt}
\begin{aligned}\label{SCA}
&{\rm S.1)}\;\;\;\widehat{\bp}[k]\;=\;\underset{\boldsymbol{p}\in \mathcal{C}}{\arg\min}\;\; \widetilde{\rm MSD}(\bp;\bp[k])\nonumber\\
&{\rm S.2)}\;\;\;\bp[k+1]\;=\;\bp[k]+\gamma[k]\Big(\widehat{\bp}[k]-\bp[k]\Big)\nonumber
\end{aligned}
\end{equation}
\end{algorithm}

\section{Recursive Least Squares Estimation \\of Graph Signals}
As is well known, LMS strategies have low complexity, but typically suffer of slow convergence rate. To improve the learning rate of the adaptive estimation task, we can employ an RLS method, which in turn has a larger computational burden, see, e.g. \cite{sayed2011adaptive}. In particular, the optimal (centralized) RLS estimate for $\bs^o$ at time $n$, say, $\widehat{\bs}_c[n]$, can be found as the vector that solves the following optimization problem:
\begin{equation}
\label{eq.rls_opt_prob}
\underset{\bs}{\text{min}}\;\,\sum_{l = 1}^{n}\beta^{n-l}\big\|\mD_{\S[l]}(\by[l] -\mU_{\F}\bs)\big\|^2_{\mathbf{C}_v^{-1}} + \beta^n\|\bs\|^2_{\boldsymbol{\Pi}}
\end{equation}
where $0 \ll \beta \leq 1$ is the exponential forgetting factor, $\boldsymbol{\Pi}\succeq \mathbf{0}$ is a regularization matrix, and we have exploited the fact that $\mD_{\S[n]}$ is an idempotent matrix for all $n$. Typically, $\boldsymbol{\Pi}=\delta \mI$, where $\delta>0$ is small \cite{sayed2011adaptive}. Solving \eqref{eq.rls_opt_prob} and using (\ref{compact_decomp}), the optimal estimate for $\bx^o$ at time $n$ is given by:
\begin{align}
\label{eq.rls_opt_sol}
\widehat{\bx}_c[n]  \,=\, \mU_{\F}\widehat{\bs}_c[n] \,=\, \mU_{\F}\bPsi^{-1}[n]\bpsi[n]
\end{align}
where
\begin{align}
&\bPsi[n]\,=\,\sum_{l= 1}^n\beta^{n-l}\mU_{\F}^H \mD_{\S[l]}\mathbf{C}_v^{-1} \mU_{\F}	 + \beta^n\boldsymbol{\Pi},\label{Psi}\\
&\bpsi[n]\,=\,\sum_{\l = 1}^{n}\beta^{n-l}\mU_{\F}^H \mD_{\S[l]}\mathbf{C}_v^{-1}\by[l]. \label{psi}
\end{align}
The regularization term in (\ref{Psi}) avoids invertibility issues in (\ref{eq.rls_opt_sol}), especially at early values of $n$. Given the structure of the recursion of $\bPsi[n]$ and $\bpsi[n]$ in \eqref{Psi}-\eqref{psi}, we obtain
\begin{align}
&\bPsi[n] \,=\, \beta\,\bPsi[n-1] + \mU_{\F}^H \mD_{\S[n]} \mathbf{C}_v^{-1}\mU_{\F}, \label{Psi_rec}\\
&\bpsi[n] \,=\, \beta\,\bpsi[n-1] + \mU_{\F}^H \mD_{\S[n]}\mathbf{C}_v^{-1}\by[n], \label{psi_rec}
\end{align}
with $\bPsi[0]=\boldsymbol{\Pi}$, which recursively update both $\bPsi[n] $ and $\bpsi[n]$ given their previous values.  Thus, the RLS algorithm for graph signals evolves as illustrated in Algorithm 5, which has computational complexity of the order of $O(|\F|^3)$, due to the presence of the inverse operation $\bPsi^{-1}[n]$. Since typically we have $|\F| \ll N$, the cost $O(|\F|^3)$ is often affordable \footnote{A sequential version of the RLS on graphs can be readily derived using the matrix inversion lemma, and leading to a complexity equal to $O(|\F|^2)$}. The properties of the RLS algorithm strongly depend on the random sampling operator $\mD_{\S[n]}$. Thus, in the sequel, we will show how the design of the random sampling strategy affects the reconstruction capability and the steady-state performance of Algorithm 5. Since the study of the mean-square performance of RLS adaptive filters is challenging \cite{sayed2011adaptive}, in the sequel we will exploit the following assumption in order to make the analysis tractable \footnote{This is a common assumption in the analysis of RLS-type algorithms, see, e.g., \cite{sayed2011adaptive}, and yields good results in practice.}.

\begin{algorithm}[t]
\caption*{\textbf{Algorithm 5: RLS on Graphs}}
\vspace{.1cm}
Start with random $\bpsi[0]$, and $\bPsi[0]=\boldsymbol{\Pi}$. For $n>0$, repeat:
\begin{equation}
\setlength{\jot}{8pt}
\begin{aligned}\label{eq.RLS}
&\bPsi[n] \,=\, \beta\,\bPsi[n-1] + \mU_{\F}^H \mD_{\S[n]}\mathbf{C}_v^{-1} \mU_{\F}\nonumber\\
&\bpsi[n] \,=\, \beta\,\bpsi[n-1] + \mU_{\F}^H \mD_{\S[n]}\mathbf{C}_v^{-1}\by[n]\\
&\widehat{\bx}_c[n]  \,=\, \mU_{\F}\bPsi^{-1}[n]\bpsi[n]
\end{aligned}
\end{equation}
\end{algorithm}

\vspace{.1cm}
\textit{Assumption 4 (Ergodicity):} $\exists n_0$ such that, for all $n>n_0$, the time average $\bPsi[n]$ in (\ref{Psi}) can be replaced by its expected value, i.e., $\bPsi[n]=\mathbb{E} \bPsi[n]$, for $n>n_0$. \qedsymbol

\subsection{Adaptive Reconstruction from Sparse Sampling}

Under Assumption 4, the steady state behavior of matrix  $\bPsi[n]$ in (\ref{Psi}) can be approximated as:
\begin{align} \label{approx}
\overline{\bPsi}\,=\,&\lim_{n\rightarrow\infty}\bPsi[n]=\, \lim_{n\rightarrow\infty}\mathbb{E}\bPsi[n]\nonumber\\
\,=\,&\frac{1}{1-\beta}\mU_{\F}^H {\rm diag}(\bp)\mathbf{C}_v^{-1} \mU_{\F}
\end{align}
where $\bp=(p_1,\ldots,p_N)^T\in[0,1]^{N}$ represents the sampling probability vector. Thus, from (\ref{eq.rls_opt_sol}) and (\ref{approx}), we deduce that asymptotic reconstruction of $\bx^o$
\textit{necessarily} requests the positive (semi-) definite matrix $\mU_{\F}^H\,{\rm diag}(\bp)\mathbf{C}_v^{-1}\mU_{\F}$ to be invertible (or full rank), which is similar to condition \eqref{lambda_min} in Sec. \ref{AdaptRecSparseSampling} for LMS reconstruction. Thus, proceeding as in Sec. \ref{AdaptRecSparseSampling}, under the assumption that the observation noise is spatially uncorrelated, i.e., matrix $\mathbf{C}_v$ is diagonal, it is easy to show that condition (\ref{|DcB|<1}) is necessary and sufficient to guarantee adaptive graph signal reconstruction using Algorithm 5.

\subsection{Mean-Square Performance}
\label{RLS_analysis}

In the sequel, we illustrate the effect of the probability vector $\bp$ on the mean-square behavior of Algorithm 5.

\vspace{-.3cm}
\textit{\begin{theorem}\label{theorem_mean_square}
Assume model (\ref{lin_observation}), condition (\ref{lambda_min}), Assumptions 2 and 4 hold. Then, Algorithm 5 is mean-square stable, with mean-square deviation given by
\begin{align}\label{MSD2}
{\rm MSD}&=\lim_{n\rightarrow\infty}\,\mathbb{E}\|\widehat{\bx}_c[n]-\bx^o\|^2  \nonumber\\
&=\,\frac{1-\beta}{1+\beta}\,{\rm Tr}\left[\left(\mU_{\F}^H\,{\rm diag}(\bp)\mathbf{C}_v^{-1}\mU_{\F}\right)^{-1}\right].
\end{align}
\end{theorem}} \smallskip

\begin{proof}
See Appendix B.
\end{proof}

\subsection{Optimal Sampling Strategies}

Exploiting the results obtained in Sec. IV-B, the goal of this section is to develop optimal sampling strategies aimed at selecting the probability vector $\bp$ to optimize the performance of the RLS on Graphs. In particular, the method leads to optimal sampling strategies aimed at selecting a probability vector $\bp$ that minimizes the total sampling rate, while guaranteing a target value of MSD. To this aim, we formulate the following optimization problem:

\begin{equation}\label{sparse_sensing}
\setlength{\jot}{6pt}
\begin{aligned}
&\min_{\boldsymbol{p}} \;\; \mathbf{1}^T\bp   \\
&\;\;\;\hbox{s.t.} \;\;\;
{\rm Tr}\left[\left(\mU_{\F}^H\,{\rm diag}(\bp)\mathbf{C}_v^{-1}\mU_{\F}\right)^{-1}\right]\leq \gamma\, \frac{1+\beta}{1-\beta} \\
&\qquad\;\;\, \;\mathbf{0}\leq \bp \leq \bp^{\max}
\end{aligned}
\end{equation}
The linear objective function in (\ref{sparse_sensing}) represents the total graph sampling rate that has to be minimized. From (\ref{MSD}), the first constraint in (\ref{sparse_sensing}) imposes that the MSD must be less than or equal to a constant $\gamma>0$ [this constraint implicitly guarantees the reconstruction condition in (\ref{lambda_min})]. As before, the last constraint limits the vector to lie in the box $p_i\in[0,p^{\max}_i]$, for all $i$. It is easy to check the convexity of problem (\ref{sparse_sensing}), whose global solution can be found using efficient algorithms \cite{boyd2004convex}. Obviously, one can consider also the related problem where we aim at minimizing the MSD in (\ref{MSD2}) while imposing a maximum budget on the total graph sampling rate, which also translates into a convex optimization program.

\subsection{Distributed Adaptive Implementation}

In many practical systems, data are collected in a distributed network, and sharing local information with a central processor might be either unfeasible or not efficient, owing to the large volume of data, time-varying network topology, and/or privacy issues \cite{cattivelli2008diffusion,mateos2009distributed}. Motivated by these observations, in this section we extend the RLS strategy in Algorithm 5 to a distributed setting, where the nodes of the graph are connected through a sparse communication network described by the graph ${\cal G}_c=(\mathcal{V},\mathcal{E}_c)$. The topology of the communication graph ${\cal G}_c$ does not necessarily coincide with that of the graph used to process the data, i.e., ${\cal G}$. To ensure the diffusion of information over the network, we assume the following.

\textit{Assumption 5 (Topology):} The communication graph $\mathcal{G}_c$ is symmetric and connected. \qedsymbol

To derive distributed solution methods for problem (\ref{eq.rls_opt_prob}), let us introduce local copies $\{\bs_i\}_{i=1}^N$ of the global variable $\bs$, and recast problem (\ref{eq.rls_opt_prob}) in the following equivalent form:
\begin{align}\label{opt_prob_dist}
&\min_{\{\boldsymbol{s}_i\}_{i=1}^N} \;\;\sum_{i=1}^N \sum_{l= 1}^{n}\beta^{n-l} \frac{d_{i}[n]}{\sigma_i^2}(y_{i}[n]-\bu_{\F,i}^H\bs_i)^2+ \frac{\beta^n}{N}\sum_{i=1}^N\|\bs_i\|^2_{\boldsymbol{\Pi}} \nonumber\\
&\qquad \hbox{subject to} \quad \bs_i=\bs_j \quad \hbox{for all $i\in \mathcal{V}$, $j\in \mathcal{N}_i$,}
\end{align}
where $\mathcal{N}_i$ denotes the set of local neighbors of agent $i$ on the communication graph ${\cal G}_c$, and $\bu_{\F,i}^H$ is the $i$-th row of matrix $\mU_{\F}$.
Now, letting $\bs=\{\bs_i\}_{i=1}^N$ and $\boldsymbol{\lambda}=\{\boldsymbol{\lambda}_{ij}\}_{i\in \mathcal{V}}^{j\in \mathcal{N}_i}$, the augmented Lagrangian for problem (\ref{opt_prob_dist}) writes as:
\begin{align}\label{Lagrangian}
\mathcal{L}_a\left(\bs,\boldsymbol{\lambda}\right) \,=\, &
\sum_{i=1}^N \sum_{l = 1}^{n}\beta^{n-l} \frac{d_{i}[l]}{\sigma_i^2}(y_{i}[l]-\bu_{\F,i}^H\bs_i)^2 \nonumber\\
&+ \frac{\beta^n}{N}\sum_{i=1}^N\|\bs_i\|^2_{\boldsymbol{\Pi}}+\sum_{i=1}^N \sum_{j\in \mathcal{N}_i}\boldsymbol{\lambda}_{ij}^T (\bs_i-\bs_j)\nonumber\\
&+\frac{\varrho}{4}\sum_{i=1}^N \sum_{j\in \mathcal{N}_i}\|\bs_i-\bs_j\|^2
\end{align}
where $\varrho>0$ is a positive coefficient. Since the augmented Lagrangian function in (\ref{Lagrangian}) is strongly convex for all $n$, we can employ ADMM to solve problem (\ref{opt_prob_dist}), see, e.g., \cite{boyd2011distributed}. The first step of ADMM aims at minimizing the augmented Lagrangian in (\ref{Lagrangian}) with respect to the primal variable $\bs$.
In particular, we apply a parallel method where each node minimizes (\ref{Lagrangian}) with respect to its local variable, while keeping fixed the variables of its neighbors at the previous iteration. The local variables at each node can then be updated as the solution of the following optimization problems:
\begin{align}\label{estimate_update2}
&\widehat{\bs}_{i}[n,k+1]\,=\, \underset{\bs_i}{\text{argmin}}\;
\sum_{l = 1}^{n}\beta^{n-l} \frac{d_{i}[l]}{\sigma_i^2} (y_{i}[l]-\bu_{\F,i}^H\bs_i)^2 \nonumber\\
&\quad+ \frac{\beta^n}{N}\|\bs_i\|^2_{\boldsymbol{\Pi}} +\sum_{j\in \mathcal{N}_i}\big(\boldsymbol{\lambda}_{ij}[n,k]-\boldsymbol{\lambda}_{ji}[n,k]\big)^T\bs_i \nonumber\\
&\quad+\frac{\varrho}{2}\sum_{j\in \mathcal{N}_i}\|\bs_i-\widehat{\bs}_j[n,k]\|^2, \quad i=1,\ldots,N.
\end{align}
Each local subproblem in (\ref{estimate_update2}) corresponds to an unconstrained minimization that admits closed-form solution given by:
\begin{align}\label{estimate_update3}
&\widehat{\bs}_{i}[n,k+1]\,=\, \Big(\bPsi_i[n] + \varrho |\mathcal{N}_i|\bI\Big)^{-1}\bigg[\bpsi_i[n]+\varrho\sum_{j\in \mathcal{N}_i}\widehat{\bs}_{j}[n,k]\nonumber\\
&\qquad-\frac{1}{2}\sum_{j\in \mathcal{N}_i} (\boldsymbol{\lambda}_{ij}[n,k]-\boldsymbol{\lambda}_{ji}[n,k])\bigg],
\end{align}
for $i=1,\ldots,N$, where, setting $\bPsi_{i}[0]=\boldsymbol{\Pi}/N$, we have
\begin{align}
&\bPsi_{i}[n] \,=\, \beta\,\bPsi_{i}[n-1] + d_{i}[n]\bu_{\F,i} \bu_{\F,i}^H/\sigma_i^2, \label{Psi_rec_loc}\\
&\bpsi_{i}[n] \,=\, \beta\,\bpsi_{i}[n-1] + d_{i}[n]y_{i}[n]\bu_{\F,i}/\sigma_i^2. \label{psi_rec_loc}
\end{align}
Finally, the second step of the ADMM algorithm updates the Lagrange multipliers as:
\begin{align}\label{mult_update}
&\boldsymbol{\lambda}_{ij}[n,k+1]=\boldsymbol{\lambda}_{ij}[n,k]+\frac{\varrho}{2}\, \Big(\widehat{\bs}_j[n,k+1]-\widehat{\bs}_i[n,k+1]\Big),
\end{align}
for $i\in \mathcal{V}$, $j\in \mathcal{N}_i$. Recursions (\ref{estimate_update3}) and (\ref{mult_update}) constitute the ADMM-based distributed RLS algorithm (DRLS), whereby all sensors $i\in \mathcal{V}$ keep track
of their local estimate $\widehat{\bs}_{i}$ and their multipliers $\{\boldsymbol{\lambda}_{ij}\}_{j\in \mathcal{N}_i}$, which can be arbitrarily initialized. Then, all the steps of the DRLS strategy for distributed adaptive reconstruction of graph signals are summarized in Algorithm 6. The proposed DRLS method has a double time-scale. The inner loop on index $k$ has the goal of forcing consensus among variables $\widehat{\bs}_{i}[n,k]$. However, when the network is deployed to track a time-varying graph signal, one cannot afford large delays in-between consecutive sensing instants. In this case, we can run a single consensus iteration per acquired observation, i.e., $K=1$ in Algorithm 6, thus making the method suitable for operation in nonstationary environments. In the sequel, we will validate the proposed distributed approach via numerical simulations.

\begin{algorithm}[t]
\caption*{\textbf{Algorithm 6: DRLS on Graphs}}
\vspace{.1cm}
Start with $\{\bpsi_{i}[0]\}_{i=1}^N$, $\{\bs_{i}[0,0]\}_{i=1}^N$, $\{\boldsymbol{\lambda}_{ij}[0,0]\}_{i\in \mathcal{V}}^{j\in \mathcal{N}_i}$ chosen at random, and set $\{\bPsi_i[0]\}_{i=1}^N=\boldsymbol{\Pi}/N$, and $\varrho\in[0,\varrho_{\max}]$. \smallskip\\ \smallskip
\textbf{for} $n> 0$ \textbf{do}\\ \smallskip
\hspace{.5cm}All $i\in \mathcal{V}$: update $\bPsi_{i}[n]$ and $\bpsi_i[n]$ using (\ref{Psi_rec_loc}) and (\ref{psi_rec_loc});\\ \smallskip
\hspace{.5cm}\textbf{for} $k=0,\ldots,K-1$ \textbf{do}\\ \smallskip
\hspace{1cm}All $i\in \mathcal{V}$: transmit $\boldsymbol{\lambda}_{ij}[n,k]$ to each $j\in \mathcal{N}_i$;\\ \smallskip
\hspace{1cm}All $i\in \mathcal{V}$: update $\widehat{\bs}_{i}[n,k+1]$ using (\ref{estimate_update3});\\ \smallskip
\hspace{1cm}All $i\in \mathcal{V}$: transmit $\widehat{\bs}_{i}[n,k+1]$ to each $j\in \mathcal{N}_i$;\\ \smallskip
\hspace{1cm}All $i\in \mathcal{V}$: update $\{\boldsymbol{\lambda}_{ij}[n,k+1]\}_{j\in \mathcal{N}_i}$ using (\ref{mult_update});\\ \smallskip
\hspace{.5cm}\textbf{end}\\ \smallskip
\textbf{end}
\end{algorithm}

\section{Numerical Results}

In the sequel, we illustrate the performance of the proposed strategies applied to both synthetic and real data.

\begin{figure*}[t]
\centering
\includegraphics[width=18cm]{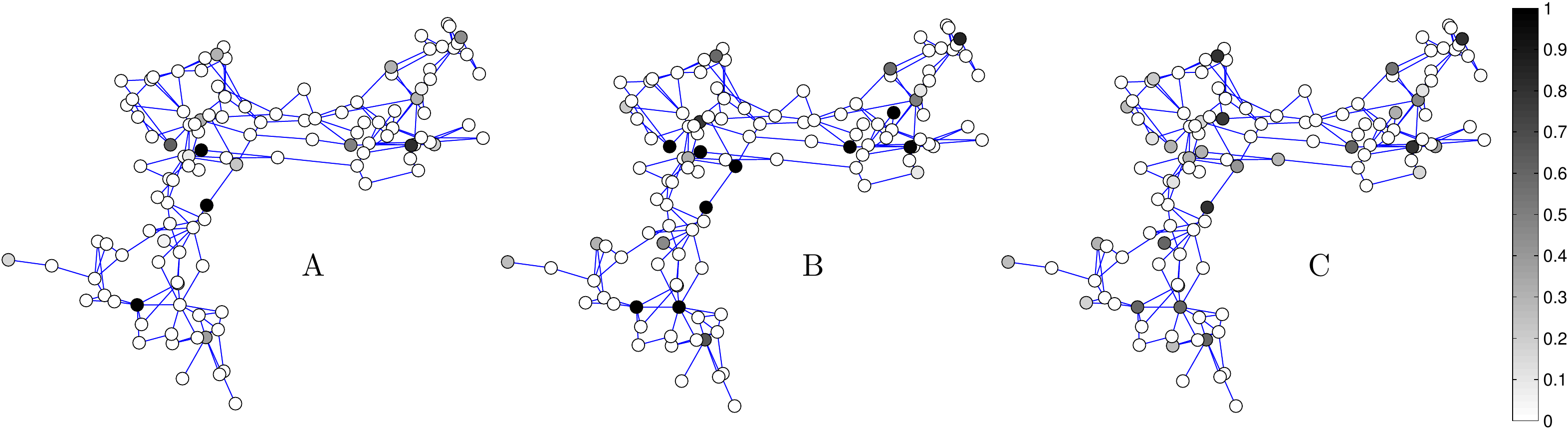}
\caption{IEEE 118 bus system: Graph Topology, and optimal probability vectors obtained from (\ref{min_MSD_problem2}), for different values of $\bar{\alpha}$.}
\label{fig:Graph}
\end{figure*}

\subsubsection{LMS on Graphs}

\begin{figure}[t]
\centering
\includegraphics[width=7.8cm]{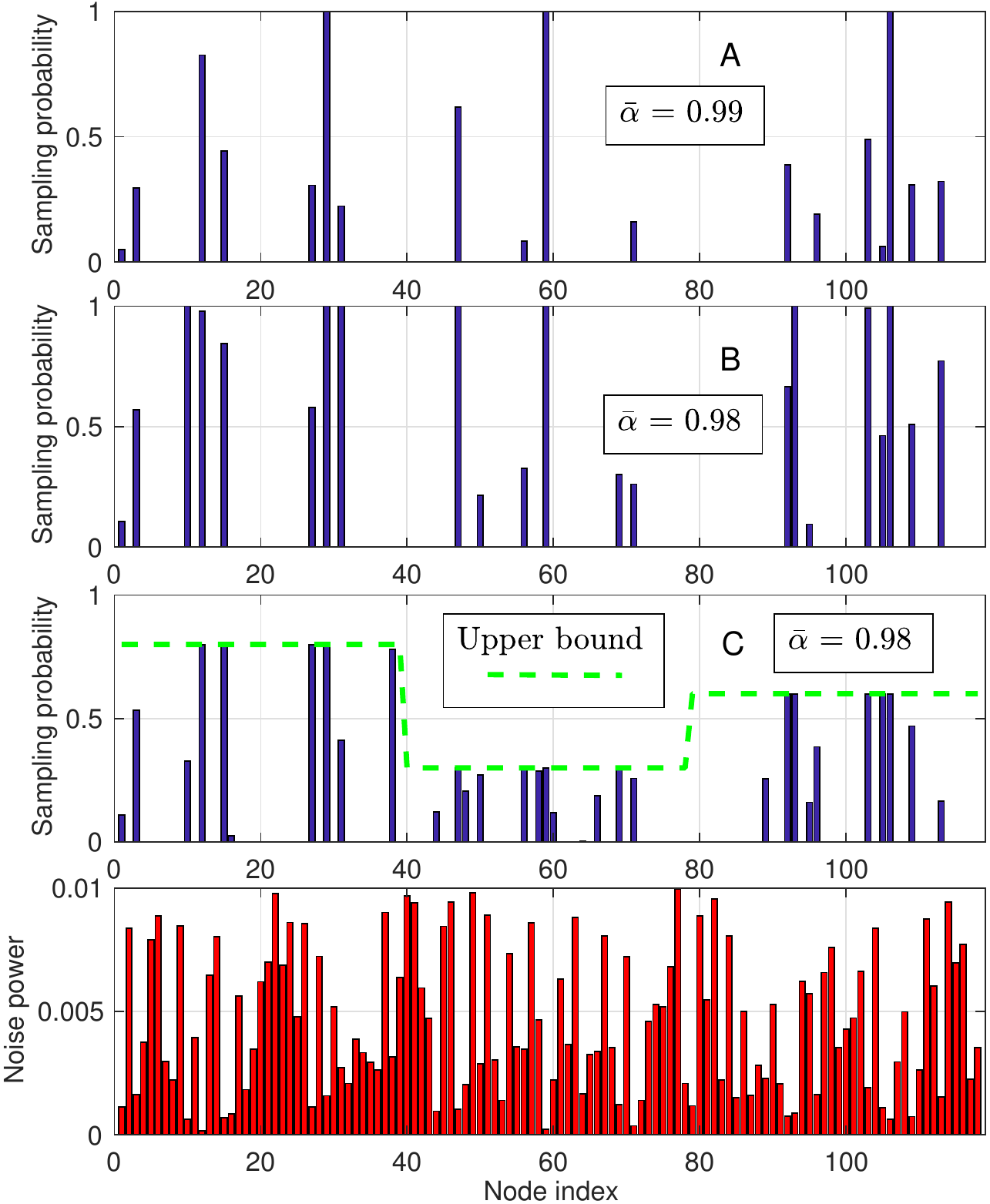}
\caption{Optimal probabilities and noise variance, obtained from (\ref{min_MSD_problem2}) with Algorithm 3 for different values of $\bar{\alpha}$ and $\bp^{\max}$.}
\label{fig:Prob}
\end{figure}

Let us consider an application to the IEEE 118 Bus Test Case, which represents a portion of the American Electric Power System (in the Mid-western US) as of December 1962. The graph is composed of 118 nodes (i.e., buses), and its topology (i.e., transmission lines connecting buses) is illustrated in Fig. \ref{fig:Graph} \cite{sun2005simulation}. The dynamics of the power generators give rise to smooth graph signals (e.g., powers, currents, voltages), whose spectral content is assumed to be limited to the first ten eigenvectors of the Laplacian matrix of the graph in Fig. \ref{fig:Graph}, i.e., $|\mathcal{F}|=10$. The observation noise in (\ref{lin_observation}) is zero-mean, Gaussian, with a diagonal covariance matrix $\bC_v$, where each element is illustrated in Fig. \ref{fig:Prob} (bottom). The other parameters are: $\mu=0.1$, and $\mathcal{P}=118$. Then, in Fig. \ref{fig:Prob} (A , B, and C), we plot the optimal probability vector obtained using Algorithm 2 for different values of $\bar{\alpha}$ (0.99 for case A, 0.98 for both cases B and C) and upper bound vectors $\bp^{\max}$ (${\bf 1}$ for cases A and B and as illustrated in the figure for case C). In all cases, we have checked that the constraint on the convergence rate in \eqref{min_MSD_problem} is attained strictly. From Fig. \ref{fig:Prob}, as expected, we notice how the method enlarges the expected sampling set if we either require (see cases A and B) a faster convergence rate (i.e., a smaller $\bar{\alpha}$), or if stricter bounds are imposed on the maximum probability to sample ``important'' nodes (see case C). Also, from Fig. \ref{fig:Prob}, it is clear that the method finds a very sparse probability vector and avoids to assign large sampling probabilities to nodes having large noise variances. We remarks that with the proposed formulation, sparse sampling patterns are obtained thanks to the optimization of the sampling probabilities, without resorting to any relaxation of complex integer optimization problems. The corresponding positions of the samples collected over the IEEE 118 Bus graph are illustrated in Fig. \ref{fig:Graph}, where the the color (in gray scale) of the vertices denotes the value of the sampling probability.

\begin{figure}[t]
\centering
\includegraphics[width=8cm]{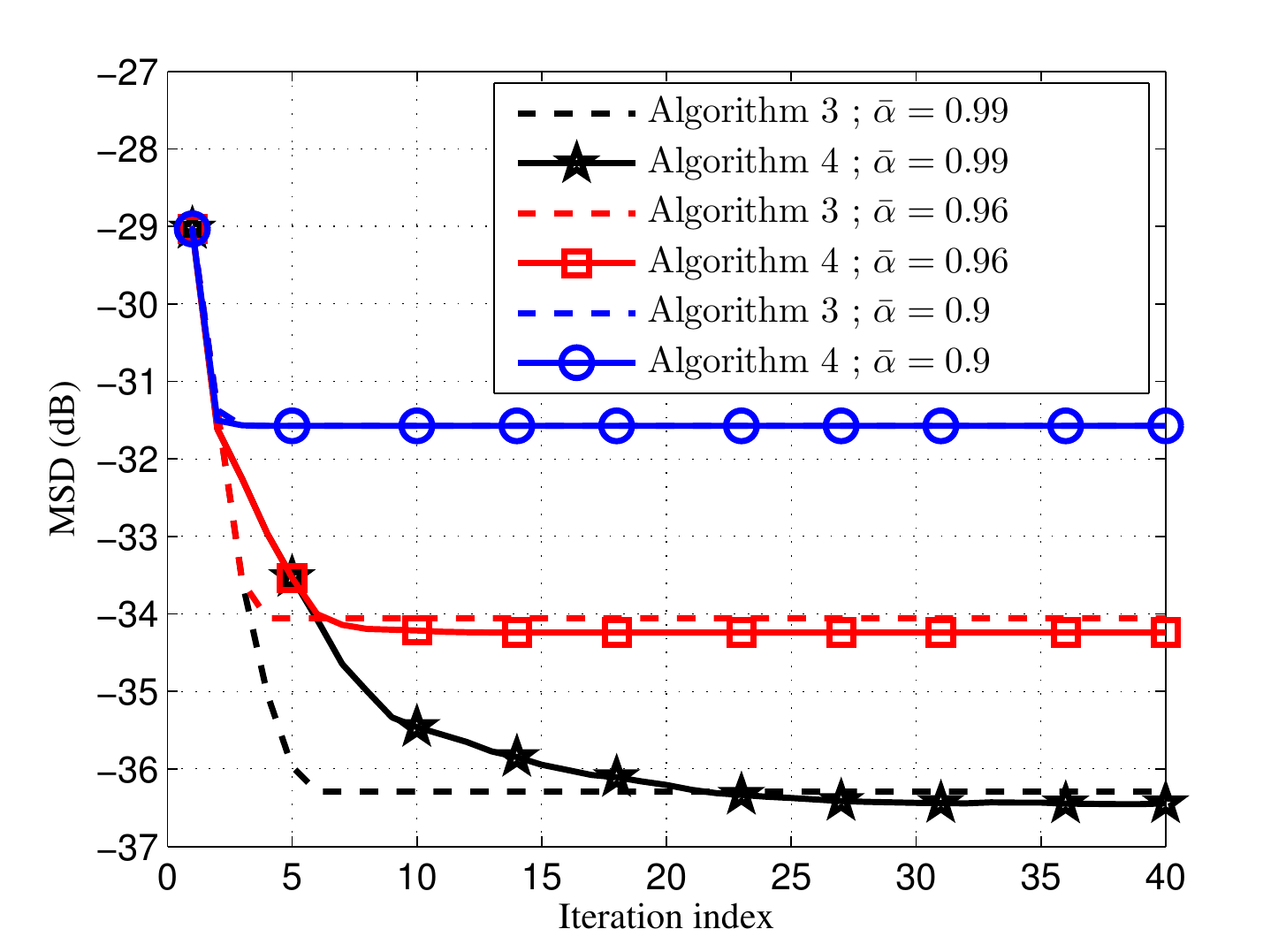}
\caption{Behavior of ${\rm MSD}(\bp[k])$ versus iteration index, obtained using Algorithms 3 and 4.}
\label{fig:MSD}
\end{figure}

To compare the results obtained using Algorithms 3 and 4, in Fig. \ref{fig:MSD}, we report the temporal evolution of ${\rm MSD}$ in (\ref{MSD}) evaluated for each instantaneous estimate $\bp[k]$, obtained using the aforementioned strategies. Both algorithms are initialized such that $p_i[1]=0.5$ for all $i$, and we set $\bp^{\max}={\bf 1}$. The SCA algorithm exploits the surrogate function in (\ref{surrogate}) with $\tau=10^{-6}$, and the step size sequence in (\ref{SCA}) obeys to the diminishing rule $\gamma[k]=\gamma[k-1](1-\eta\gamma[k-1])$, with $\gamma[0]=1$, and $\eta=0.001$. The surrogate problem in S.1 of Algorithm 4 is solved using the CVX software \cite{boyd2004convex}. As we can see from Fig.~\ref{fig:MSD}, both algorithms illustrate a very good convergence behavior, reaching their final state in a few iterations. In particular, we notice that the Algorithm 3 is slightly faster than Algorithm 4. But most importantly, we notice that both methods converge to very similar final solutions. This means that local optimal solutions of the original non-convex problem (\ref{min_MSD_problem}) are very similar to global optimal solutions of the approximated convex/concave fractional problem in (\ref{min_MSD_problem2}). This result justifies the approximation made in (\ref{MSD_bound}) to formulate the sampling design problem (\ref{min_MSD_problem2}) and to derive Algorithm 3, which also guarantees convergence to global optimal solutions.

\begin{figure}[t]
\centering
\includegraphics[width=8.2cm]{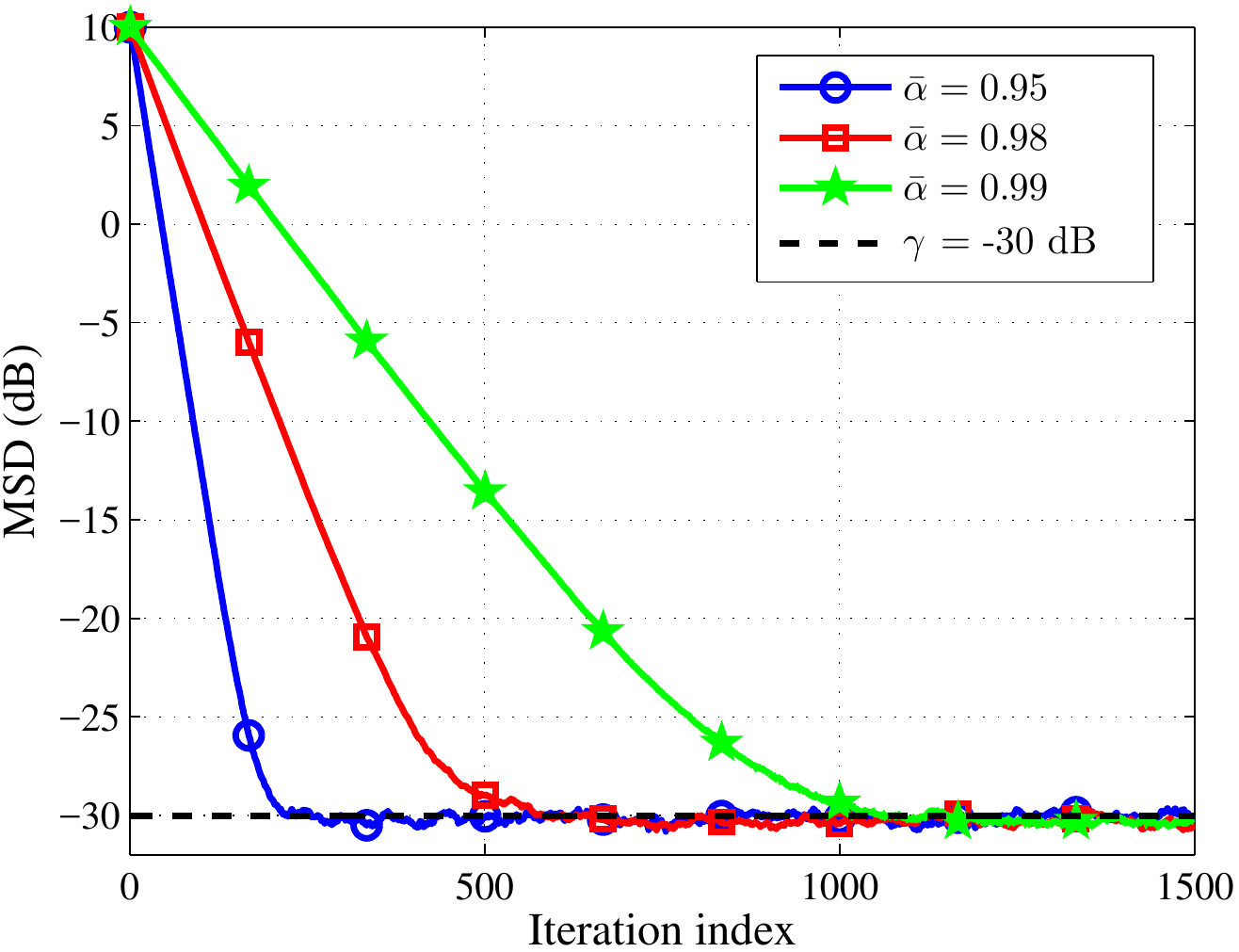}
\caption{Learning curve of Algorithm 1, obtained solving (\ref{min_sampling_rate_problem2}) for different values of $\bar{\alpha}$.}
\label{fig:MSD_alpha}
\end{figure}

To validate the theoretical results derived in Sec. \ref{LMS_analysis}, in Fig. \ref{fig:MSD_alpha} we report the learning curve (in terms of MSD) of the LMS on Graphs (cf. Algorithm 1), obtained solving (\ref{min_sampling_rate_problem2}) for different values of $\bar{\alpha}$. In particular, we set a target performance on the MSD given by $\gamma=-30$ dB. The other parameters are: $\mu=0.1$, $\bp^{\max}=\mathbf{1}$, and $\mathcal{P}=118$. The curves are averaged over 100 independent simulations. As we can see from Fig. \ref{fig:MSD_alpha}, thanks to the effect of the sampling strategy (\ref{min_sampling_rate_problem2}), the LMS algorithm can increase its convergence rate (reducing the value of $\bar{\alpha}$), while always guaranteing the performance requirement on the MSD. This curve also confirms the theoretical analysis derived in Sec. \ref{LMS_analysis}, and further justifies the approximation of the MSD function made in (\ref{MSD_bound}).

\begin{figure}[t]
\centering
\includegraphics[width=8.2cm]{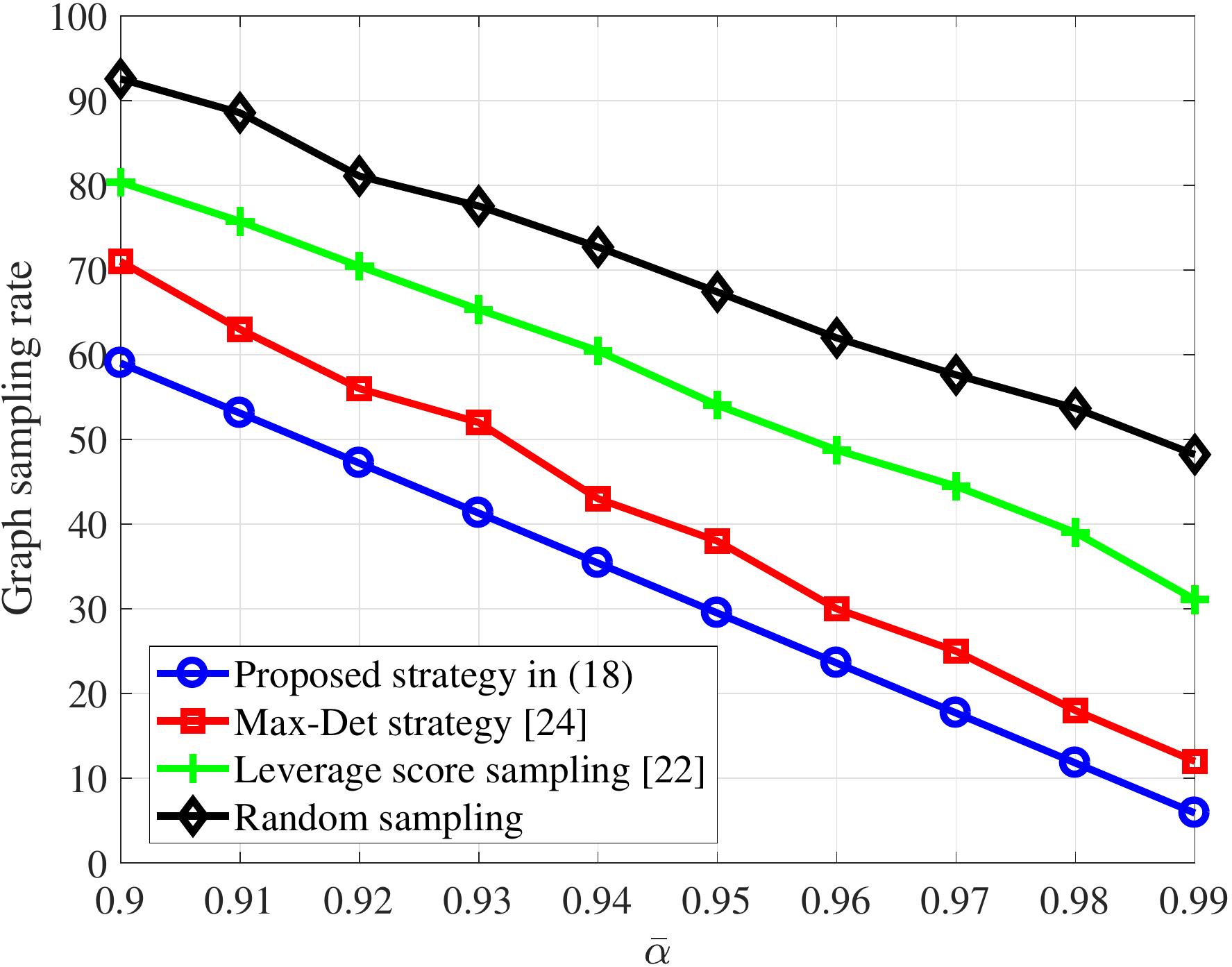}
\caption{Graph sampling rate versus convergence rate $\bar{\alpha}$, for different sampling strategies.}
\label{fig:comparisons}
\end{figure}

Finally, we compare the sampling strategy in (\ref{min_sampling_rate_problem2}) with some established sampling methods for graph signals, namely, the leverage score sampling from \cite{chen2015signalrecovery}, the Max-Det greedy strategy from \cite{tsitsvero2015signals}, and the (uniformly) random sampling. For each strategy, we keep adding nodes to the sampling set according to the corresponding criterion until the constraints on the convergence rate and the MSD in (\ref{min_sampling_rate_problem2}) are satisfied. Then, in Fig. \ref{fig:comparisons}, we report the behavior of the total graph sampling rate $\mathbf{1}^T\bp$ versus the parameter $\bar{\alpha}$ in (\ref{min_sampling_rate_problem2}), obtained using the four aforementioned strategies. The other parameters are: $\mu=0.1$, $\bp^{\max}=\mathbf{1}$, and $\gamma=-25$ dB. The results for the random sampling strategies are averaged over 200 independent simulations. As expected, from Fig. \ref{fig:comparisons}, we notice how the graph sampling rate increases for lower values of $\bar{\alpha}$, i.e., increasing the convergence rate of the algorithm, for all strategies. Furthermore, we can notice the large gain
in terms of reduced sampling rate $\mathbf{1}^T \bp$ obtained by the proposed strategy with respect to other methods available in the literature.

\subsubsection{RLS on Graphs}

\begin{figure}[t]
\centering
\includegraphics[width=7.8cm]{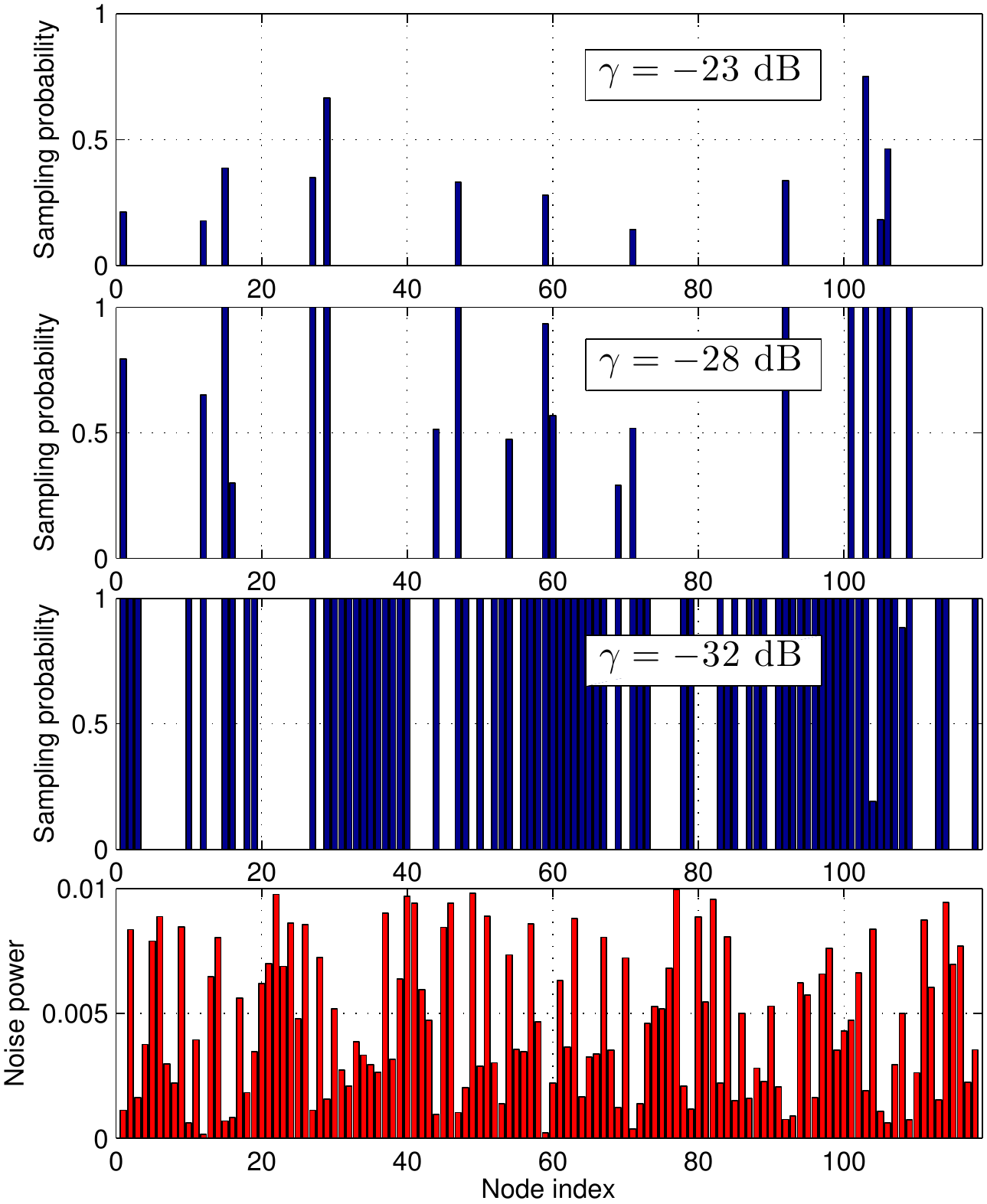}
\caption{Optimal sampling probability vector, and noise variance, versus node index, for different values of $\gamma$.}
\label{fig:Prob_RLS}
\end{figure}

Let us consider the same setting as before. An example of optimal probabilistic sampling, obtained solving Problem (\ref{sparse_sensing}), with $\beta=0.95$, is illustrated in Fig. \ref{fig:Prob_RLS}, for three different values of $\gamma$. As we can notice from Fig. \ref{fig:Prob_RLS}, the method finds a very sparse probability vector in order to guarantee the performance requirement on the MSD. In all cases, we have checked that the constraint on the MSD is attained strictly. As expected, from Fig. \ref{fig:Prob_RLS}, we notice how the proposed method enlarges the expected sampling set if we require a stricter requirement on the MSD, i.e., a lower value of $\gamma$, and avoids to select nodes having large noise variance (at least at low values of $\gamma$).

\begin{figure}[t]
\centering
\includegraphics[width=8cm]{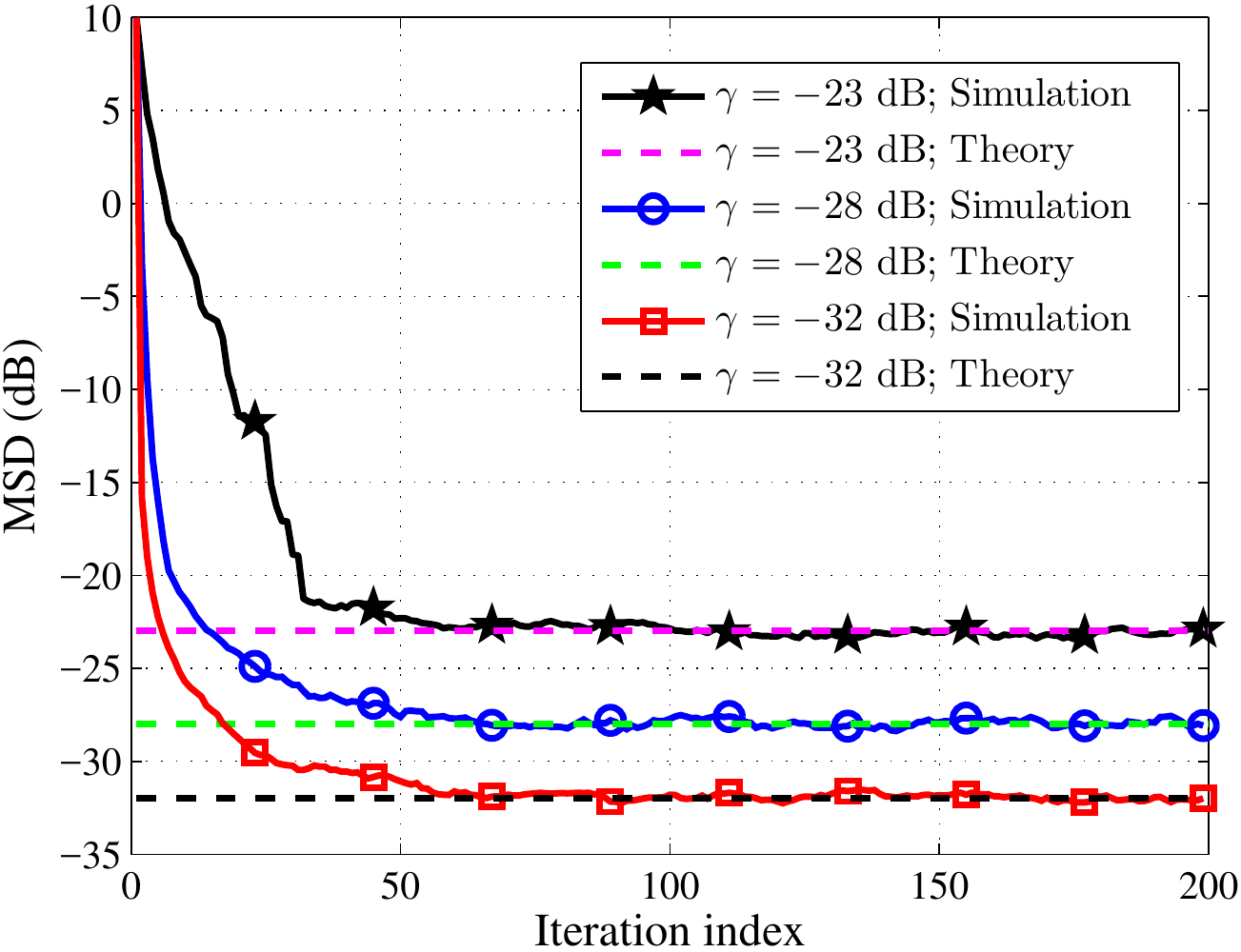}
\caption{MSD versus iteration index $n$, obtained using Algorithm 5 and the Sampling Strategy in (\ref{sparse_sensing}).}
\label{fig:MSD_RLS}
\end{figure}
\begin{figure}[t]
\centering
\includegraphics[width=8cm]{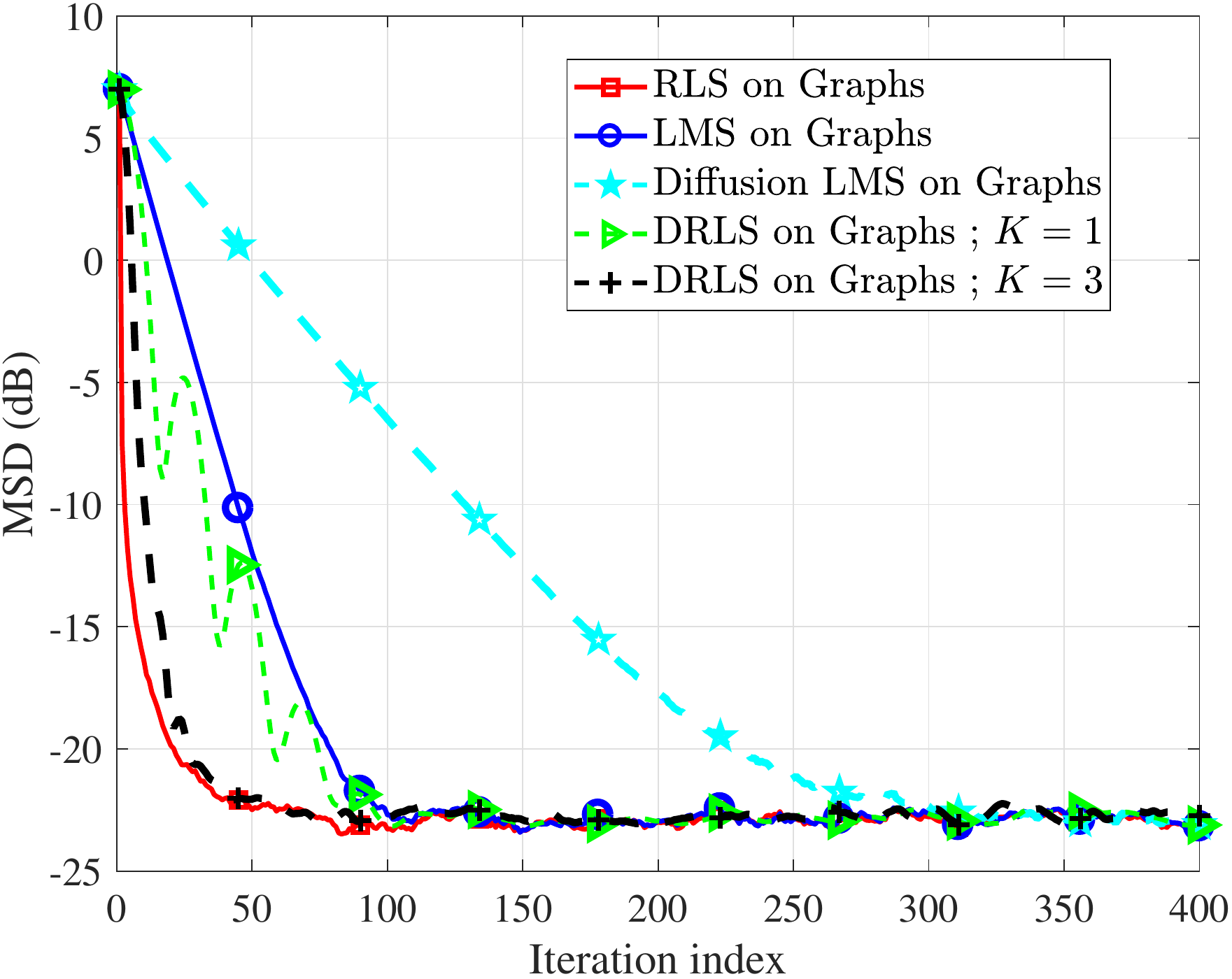}
\caption{${\rm MSD}$ versus iteration index $n$, for different algorithms.}
\label{fig:MSD_comp}
\end{figure}

To validate the theoretical results derived in Sec. \ref{RLS_analysis}, in Fig. \ref{fig:MSD_RLS} we report the learning curve (in terms of MSD) of the RLS on graphs (cf. Algorithm 5), with probability vector $\bp$ obtained by solving Problem (\ref{sparse_sensing}) for three different values of $\gamma$ (the same as in the previous example). The curves are averaged over 200 independent simulations. The theoretical expression of MSD in (\ref{MSD2}) is also reported for comparison purposes, for all values of $\gamma$. As we can see from Fig. \ref{fig:MSD_RLS}, simulations match well the theoretical results.

Finally, in Fig. \ref{fig:MSD_comp}, we report the temporal evolution of the ${\rm MSD}$ obtained using four different algorithms: the proposed LMS on Graphs in Algorithm 1; the RLS on Graphs in Algorithm 5; the Diffusion LMS on Graphs from \cite{di2016distAdaGraph}; and the DRLS strategy in Algorithm 6, considering different numbers $K$ of inner consensus iterations, i.e., $K=1$ and $K=3$. We consider a graph composed of $N=20$ nodes, whose topology is obtained from a random geometric graph model, and having algebraic connectivity equal to 0.82. The graph signal is such that $|\F|=5$. The sampling strategy of the RLS (and of the DRLS) was designed solving problem (\ref{sparse_sensing}), where $\gamma=-23$ dB, $\beta=0.95$, and $\bp^{\max}=\mathbf{1}$. The sampling strategy of the LMS (and of the Diffusion LMS) was obtained solving problem (\ref{min_MSD_problem2}), with $\mathcal{P}$ given by the optimal sampling rate obtained by the RLS strategy in (\ref{sparse_sensing}), $\bp^{\max}=\mathbf{1}$, $\mu=0.1$, and $\bar{\alpha}$ chosen to match the MSD steady-state performance of the RLS method. The communication graph is chosen equal to the processing graph. As we can see from Fig. \ref{fig:MSD_comp}, the RLS strategy is faster than the LMS on Graphs, at the cost of a higher complexity. Also, the proposed DRLS significantly outperforms the Diffusion LMS on graphs proposed in \cite{di2016distAdaGraph}. Finally, increasing the number $K$ of inner consensus iterations, the behavior of the DRLS algorithm approaches the performance of the centralized RLS, at the cost of a larger number of exchanged parameters over the network.

\begin{figure}[t]
\centering
\includegraphics[width=8cm]{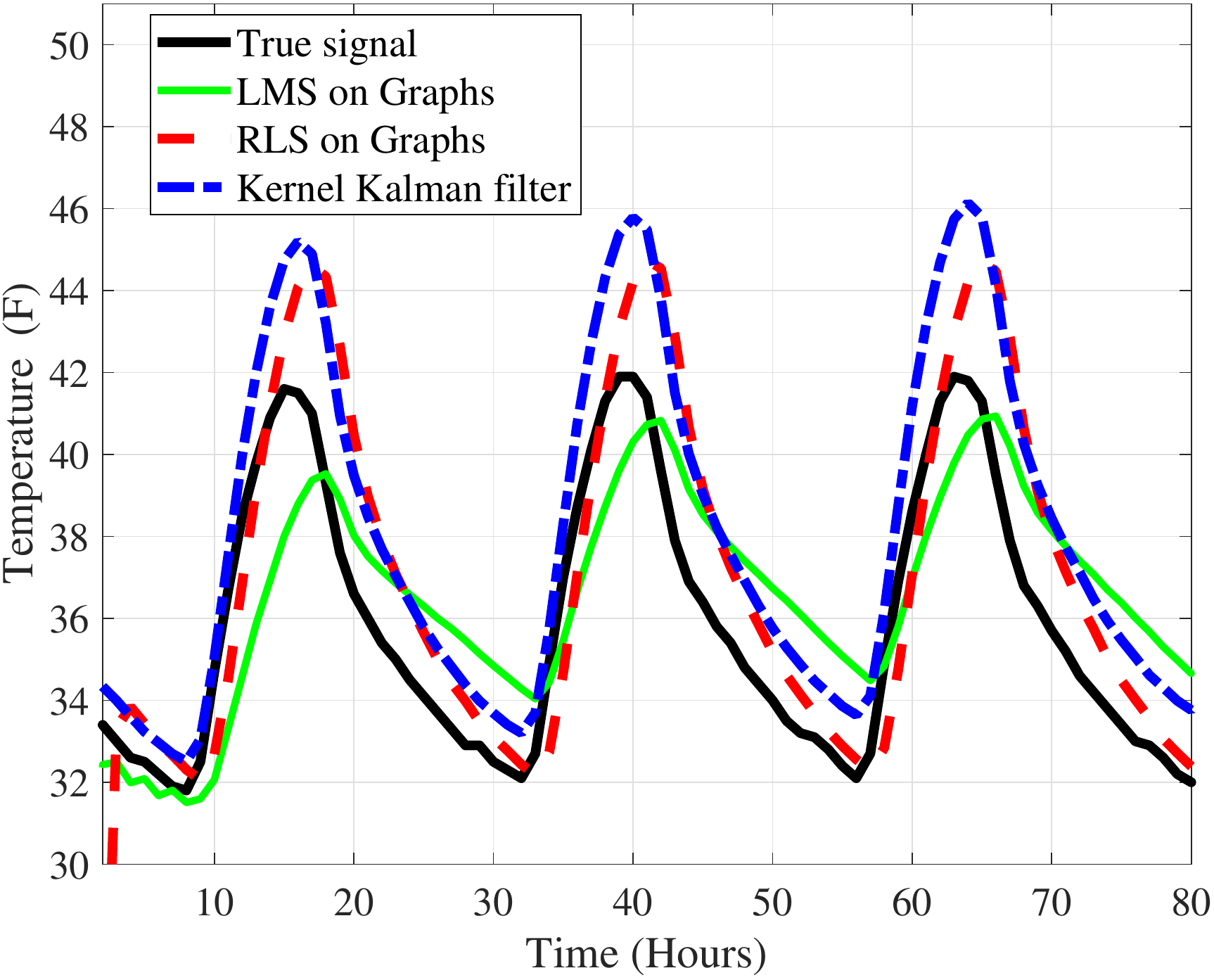}
\caption{True temperature and estimates across time at a randomly picked unobserved station.}
\label{fig:track_temp}
\end{figure}

\begin{figure}[t]
\centering
\includegraphics[width=8cm]{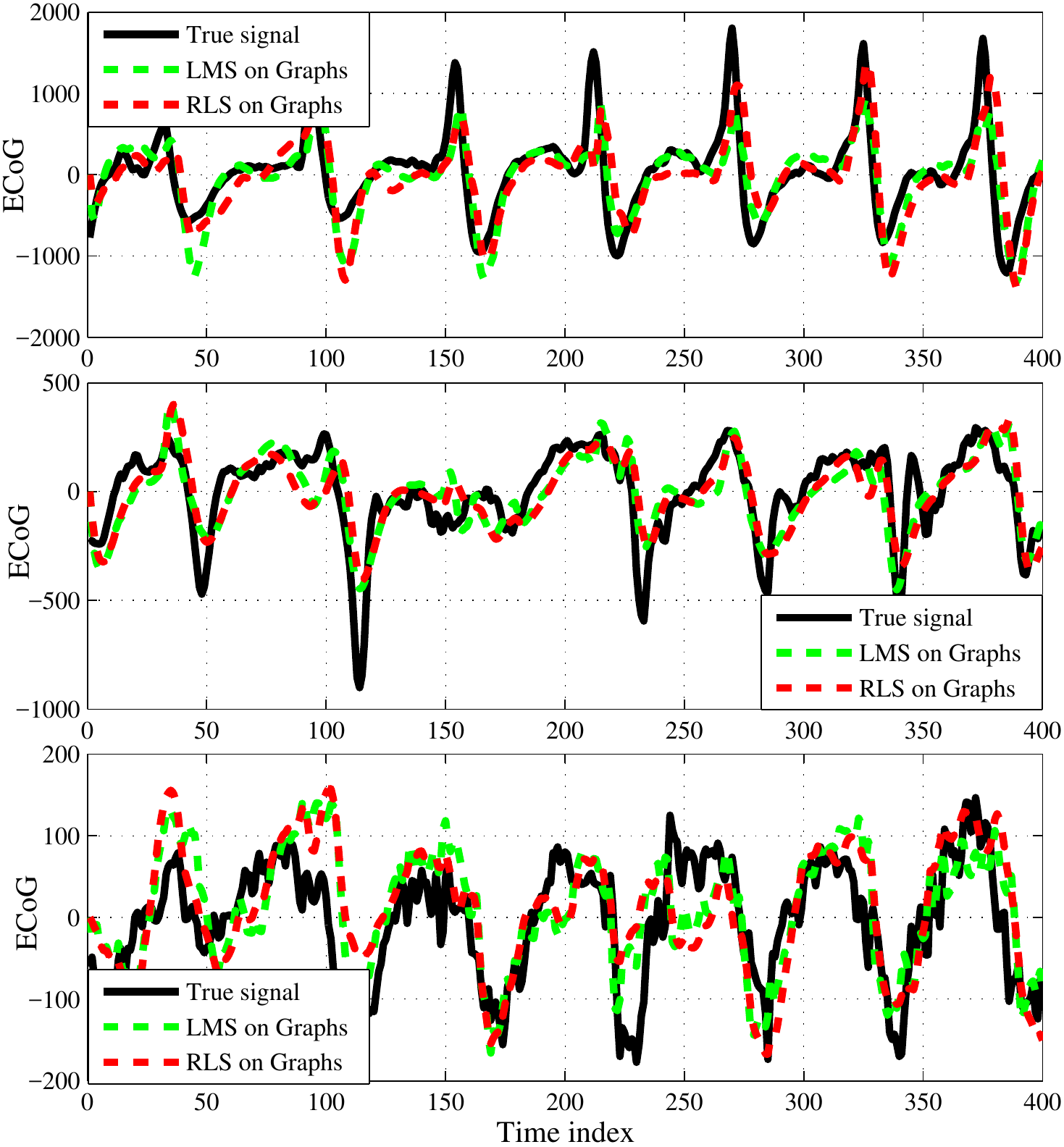}
\caption{True ECoG and estimates across time at three randomly selected unobserved regions of the brain.}
\label{fig:track_brain}
\end{figure}

\subsubsection{Tracking of Real Data}

The first data set collects temperature measurements at $N=109$ stations across the United States in 2010  \cite{temperature_dataset}. A similarity  graph  was  built using a 7 nearest neighbors approach \cite{romero2016kernel}, which relies on geographical distances. The graph signal at each vertex $i$ represents the temperature observed at the $i$-th  station and $n$-th  sampling instant. In Fig. \ref{fig:track_temp}, we illustrate the true behavior of the temperature measured at an unobserved station chosen at random, over the first 80 hours of 2010, along with its estimates carried out using three different algorithms: the LMS on Graphs in Algorithm 1 (with $\mu=1.5$ and $|\F|=40$), the RLS on Graphs (with $\beta=0.5$ and $|\F|=40$) in Algorithm 5, and the Kernel Kalman Filter (KKF) from \cite{romero2016kernel} (using the same settings as in \cite{romero2016kernel}). Since in this experiment we do not have information about observation noise, the sampling set is chosen equal for all algorithms and fixed over time, according to the Max-Det strategy in \cite{tsitsvero2015signals}, selecting $50$ samples. As we can notice from Fig. \ref{fig:track_temp}, the proposed LMS and RLS on graphs show good tracking performance that are comparable with KKF, while having a lower computational complexity.

The second example  presents  test  results  on  Electrocorticography (ECoG) data,  captured through  experiments  conducted  in  an epilepsy study \cite{kramer2008emergent}. Data were collected over a  period  of  five days, where the electrodes recorded 76 ECoG time series, consisting of voltage levels measured in different regions of the brain (see \cite{kramer2008emergent} for further details).
The GFT matrix $\mU_{\F}$ in (\ref{compact_decomp}) is learnt from the first 200 samples of ictal data, using the method proposed in \cite{gavish2017optimal}, and imposing a bandwidth equal to $|\F|=30$. In Fig. \ref{fig:track_brain}, we illustrate the true behavior of the ECoG present at three unobserved electrodes chosen at random, over the first 400 samples of ictal data, along with estimates carried out using two different algorithms: the LMS on Graphs in Algorithm 1 (with $\mu=1.5$), and the RLS on Graphs (with $\beta=0.7$)  in Algorithm 5. As before, the sampling set is chosen equal for all algorithms and fixed over time, according to the Max-Det strategy in \cite{tsitsvero2015signals}, selecting $32$ samples over the graph. As we can notice from Fig. \ref{fig:track_brain}, both methods are capable to efficiently infer and track the unknown dynamics of ECoG data at unobserved regions of the brain.

\section{Conclusions}

In this paper, we have introduced LMS and RLS strategies for adaptive learning of graph signals based on a probabilistic sampling mechanism over the graph. A mean-square analysis sheds light on how the random sampling strategy affects the performance of the proposed methods, and paved the way to the formulation of several criteria aimed at designing the sampling probabilities as an optimal trade-off between graph sampling rate, mean-square performance, and learning rate of the algorithms. Finally, a distributed RLS strategy is derived and is shown to be convergent to its centralized counterpart. Numerical simulations carried out on both synthetic and real data illustrate the good performance of the proposed methods for (possibly distributed) adaptive learning of graph signals.

As a final comment, we would like to remark that the proposed methods can be applied to the adaptive processing of signals residing on an arbitrary subspace not necessarily related to a graph. In this sense, the methods have a broader applicability that is not strictly limited to GSP. Nevertheless, the sampling operation, which lies at the core of this paper, is the aspect that finds a more direct explanation if we think at the useful signal as residing over a graph, the graph that makes that signal be represented as bandlimited.

\appendices

\section{Proof of Theorem 1}

Following energy conservation arguments \cite{sayed2003energy}, we consider a general weighted squared error sequence $\|\widetilde{\bs}[n]\|^2_{\boldsymbol{\Phi}}=\widetilde{\bs}[n]^H\boldsymbol{\Phi}\widetilde{\bs}[n]$, where $\boldsymbol{\Phi}\in \mathbb{C}^{|\F|\times|\F|}$ is an Hermitian nonnegative-definite matrix that we are free to choose.
Of course, since $\|\widetilde{\bs}[n]\|^2=\|\widetilde{\bx}[n]\|^2$ for all $n$ [cf. (\ref{compact_decomp})], it is equivalent to consider the mean-square behavior of $\widetilde{\bs}[n]$ or $\widetilde{\bx}[n]$. Then, from (\ref{error_recursion}), exploiting Assumption 2 and the fact that $\mathbb{E}\{\mD_{\S[n]}\bv[n]\bv[n]^H\mD_{\S[n]}\}=\mP\mC_v$, we can establish:
\begin{align}\label{var_relation}
\hspace{-.1cm}\mathbb{E}\|\widetilde{\bs}[n+1]\|^2_{\boldsymbol{\Phi}}= \mathbb{E}\|\widetilde{\bs}[n]\|^2_{\boldsymbol{\Phi}'}+\mu^2 \,{\rm Tr}(\boldsymbol{\Phi}\mU_{\F}^H\mP\mC_v\mU_{\F})
\end{align}
where ${\rm Tr}(\cdot)$ denotes the trace operator, matrix $\mP={\rm diag}(\bp)=\mathbb{E}\{\mD_{\S[n]}\}$, and
\begin{align}
\hspace{-.28cm}\boldsymbol{\Phi}'&\,=\,\mathbb{E} \left(\mI-\mu\,\mU_{\F}^H\mD_{\S[n]}\mU_{\F}\right)\boldsymbol{\Phi}\left(\mI-\mu\,\mU_{\F}^H\mD_{\S[n]}\mU_{\F}\right)\label{Phi'}\\
&\,=\, \boldsymbol{\Phi} -\mu\,\boldsymbol{\Phi}\mU_{\F}^H\mP\mU_{\F}-\mu\,\mU_{\F}^H\mP\mU_{\F}\boldsymbol{\Phi}\nonumber\\
&\qquad+\mu^2\mathbb{E}\{\mU_{\F}^H\mD_{\S[n]}\mU_{\F}\boldsymbol{\Phi}\mU_{\F}^H\mD_{\S[n]}\mU_{\F}\}.\label{Phi2}
\end{align}
The last term in (\ref{Phi2}) can be computed in closed form. Letting
$\mU_{\F}\mD_{\S[n]}\mU_{\F}^H\,=\,\sum_{i=1}^N d_i[n] \bu_{\F,i}\bu_{\F,i}^H$,
where $\bu_{\F,i}^H$ is the $i$-th row of $\mU_{\F}$, the last term in (\ref{Phi2}) is given by:
\begin{align}\label{last_term}
\mathbb{E}\{\mU_{\F}^H\mD_{\S[n]}\mU_{\F}\boldsymbol{\Phi}\mU_{\F}^H\mD_{\S[n]}\mU_{\F}\}= \sum_{i=1}^N \sum_{j=1}^N m_{i,j}^{(2)}\mC_{ij}
\end{align}
where $\mC_{ij}= \bu_{\F,i}\bu_{\F,i}^H \,\boldsymbol{\Phi}\,\bu_{\F,j}\bu_{\F,j}^H$, and
\begin{align}\label{Block3}
\quad m^{(2)}_{i,j}=\mathbb{E}\{d_i[n]d_j[n]\}=    \begin{cases}
      p_i, & \text{if}\;\; i=j; \\
      p_ip_j, & \text{if}\;\; i\neq j.
    \end{cases}
\end{align}
In the sequel, to study mean-square stability of Algorithm 1, we consider the following approximation:
\begin{align}
\hspace{-.28cm}\boldsymbol{\Phi}'&\,=\,\mathbb{E} \left(\mI-\mu\,\mU_{\F}^H\mD_{\S[n]}\mU_{\F}\right)\boldsymbol{\Phi}\left(\mI-\mu\,\mU_{\F}^H\mD_{\S[n]}\mU_{\F}\right)\\
&\,\simeq\,\left(\mI-\mu\,\mU_{\F}^H\mP\mU_{\F}\right)\boldsymbol{\Phi}\left(\mI-\mu\,\mU_{\F}^H\mP\mU_{\F}\right),\label{Phi_approx}
\end{align}
which is accurate under Assumption 3, i.e., for small step-sizes\footnote{This kind of approximations are typical when studying long-term dynamics of adaptive filters, see, e.g., \cite[Ch.4]{sayed2014adaptation}, and lead to good results in practice.}. In particular, it is immediate to see that (\ref{Phi_approx}) can be obtained from (\ref{Phi2})-(\ref{last_term}), by substituting the terms $p_i$ in (\ref{Block3}) with $p_i^2$, for the case $i=j$. Such approximation appears in (\ref{Phi_approx}) only in the term $O(\mu^2)$ and, under Assumption 3, it is assumed to produce a negligible deviation from (\ref{Phi'}).

Now, we proceed by showing the stability conditions for recursion (\ref{var_relation}). Letting $\mH=\mU_{\F}^H\mP\mU_{\F}$, $\mQ=(\mI-\mu\mH)^2$, and $\boldsymbol{\Phi}=\mI$, recursion (\ref{var_relation}) can be bounded as:
\begin{align}\label{bound}
\mathbb{E}\|\widetilde{\bs}[n]\|^2\leq  \mathbb{E}\|\widetilde{\bs}[0]\|^2_{\mathbf{Q}^{n} }
+\mu^2 c \sum_{l=0}^n\left\|\mQ\right\|^l
\end{align}
where $c=\,{\rm Tr}(\mU_{\F}^H\mP\mC_v\mU_{\F})$. We also have
\begin{align}\label{bound_norm_Q}
\|\mQ\|&=\|\mI-\mu\mU_{\F}^H\mP\mU_{\F}\|^2=\left(\rho\big(\mI-\mu\mU_{\F}^H\mP\mU_{\F}\big)\right)^2\nonumber\\
&\;\leq \;\max\left\{(1-\mu\delta)^2,(1-\mu\nu)^2\right\} \nonumber\\
&\;\stackrel{(a)}{\leq} \;1-2\mu\nu+\mu^2\delta^2 =1-2\mu\nu \left(1-\frac{\mu}{2 \nu}\delta^2\right)
\end{align}
where $\rho(\mX)$ denotes the spectral radius of matrix $\mX$, $\delta=\lambda_{\max}(\mU_{\F}^H\mP\mU_{\F})$, $\nu=\lambda_{\min}(\mU_{\F}^H\mP\mU_{\F})$, and in (a) we have exploited $\delta\geq\nu$. Taking the limit of (\ref{bound}) as $n\rightarrow\infty$, and since $\|\mQ\|<1$ if conditions (\ref{lambda_min}) [i.e., (\ref{|DcB|<1})] and (\ref{step}) hold, we obtain
\begin{align}\label{bound2}
\lim_{n\rightarrow\infty}\mathbb{E}\|\widetilde{\bs}[n]\|^2\leq\frac{\mu^2 c}{1-\|\mQ\|}\stackrel{(a)}{\leq}\frac{\mu c}{2\nu-\mu\delta^2},
\end{align}
where (a) follows from (\ref{bound_norm_Q}). The upper bound (\ref{bound2}) does not exceed $\mu c/\nu$ if $0<\mu<\nu/\delta^2$. Thus, for sufficiently small values of the step-size, i.e., under Assumption 3, it holds
\begin{equation}\label{lim_sup}
\lim_{n\rightarrow\infty}\,\mathbb{E}\|\widetilde{\bs}[n]\|^2 \,=\, O(\mu).
\end{equation}
Also, from (\ref{bound}), the transient component of $\mathbb{E}\|\widetilde{\bs}[n]\|^2$ vanishes as $\mQ^n$, for $n\rightarrow\infty$. Thus, from (\ref{bound_norm_Q}), the convergence rate of Algorithm 1 (i.e., $\|\mQ\|$) is well approximated by (\ref{learning_rate}) when $\mu \ll 2\nu/\delta^2$. Finally, to derive (\ref{MSD}), we recast (\ref{var_relation}) as:
\begin{align}\label{var_relation2}
&\mathbb{E}\|\widetilde{\bs}[n+1]\|^2_{\boldsymbol{\Phi}}= \mathbb{E}\|\widetilde{\bs}[n]\|^2_{\boldsymbol{\Phi}}
-\mu \mathbb{E}\|\widetilde{\bs}[n]\|^2_{\mathbf{H}\boldsymbol{\Phi}+\boldsymbol{\Phi}\mathbf{H}}
\nonumber\\
&\qquad+\mu^2 \mathbb{E}\|\widetilde{\bs}[n]\|^2_{\mathbf{H}\boldsymbol{\Phi}\mathbf{H}}+\mu^2 \,{\rm Tr}(\boldsymbol{\Phi}\mU_{\F}^H\mP\mC_v\mU_{\F})
\end{align}
Taking the limit of (\ref{var_relation2}) as $n\rightarrow\infty$ (assuming that convergence conditions are satisfied), since $\displaystyle\lim_{n\rightarrow\infty}\,\mathbb{E}\|\widetilde{\bs}[n+1]\|^2=\lim_{n\rightarrow\infty}\,\mathbb{E}\|\widetilde{\bs}[n]\|^2$ and $\displaystyle\lim_{n\rightarrow\infty}\mathbb{E}\|\widetilde{\bs}[n]\|^2_{\mathbf{H}\boldsymbol{\Phi}\mathbf{H}}=O(\mu)$ [cf. (\ref{lim_sup})], we obtain the following expression:
\begin{align}\label{var_relation3}
\hspace{-.2cm}\lim_{n\rightarrow\infty} \mathbb{E}\|\widetilde{\bs}[n]\|^2_{\mathbf{H}\boldsymbol{\Phi}+\boldsymbol{\Phi}\mathbf{H}}
=\mu \,{\rm Tr}(\boldsymbol{\Phi}\mU_{\F}^H\mP\mC_v\mU_{\F})+O(\mu^2)
\end{align}
Using now $\boldsymbol{\Phi}=\mH^{-1}$ in (\ref{var_relation3}), we obtain (\ref{MSD}).

\section{Proof of Theorem 2}
From (\ref{eq.RLS}), (\ref{Psi}), and (\ref{psi}), we obtain:
\begin{align}\label{RLS_behavior}
&\widehat{\bx}_c[n]=\mU_{\F}\left(\sum_{l = 1}^n\beta^{n-l}\mU_{\F}^H \mD_{\S[l]}\mathbf{C}_v^{-1} \mU_{\F}	 + \beta^n\boldsymbol{\Pi}\right)^{-1}\times \nonumber\\
&\qquad\qquad\times\left(\sum_{l= 1}^n\beta^{n-l}\mU_{\F}^H \mD_{\S[l]}\mathbf{C}_v^{-1}\by[l]\right)
\end{align}
Exploiting now Assumption 4, the relations (\ref{lin_observation}), (\ref{approx}), and (\ref{|DcB|<1}) [i.e., (\ref{lambda_min})], for sufficiently large $n$, the long-term behavior of recursion (\ref{RLS_behavior}) is well approximated by:
\begin{align}\label{RLS_behavior2}
&\widehat{\bx}_c[n]=\bx^o+\mU_{\F}\overline{\boldsymbol{\Psi}}^{-1}\sum_{l = 1}^n\beta^{n-l}\mU_{\F}^H \mD_{\S[l]}\mathbf{C}_v^{-1}\bv[l],
\end{align}
where we have neglected the term $\beta^n\boldsymbol{\Pi}$ at large values of $n$. Thus, from (\ref{RLS_behavior2}), we have
\begin{align}\label{error}
&\lim_{n\rightarrow\infty}\mathbb{E}\|\widehat{\bx}_c[n]-\bx^o\|^2\nonumber\\
&\hspace{-.3cm}\stackrel{(a)}{=} \hspace{-.1cm}\lim_{n\rightarrow\infty} \mathbb{E} \hspace{-.1cm} \sum_{l= 1}^n\beta^{2(n-l)} \bv^H[l]\mathbf{C}_v^{-1}\mD_{\S[l]}\mU_{\F}\overline{\boldsymbol{\Psi}}^{-2}\mU_{\F}^H \mD_{\S[l]}\mathbf{C}_v^{-1}\bv[l]\nonumber\\
&\stackrel{(b)}{=}\frac{1}{1-\beta^2}{\rm Tr} \left(\mU_{\F}^H {\rm diag}(\bp)\mathbf{C}_v^{-1}\mU_{\F}\overline{\boldsymbol{\Psi}}^{-2}\right)\nonumber\\
&\stackrel{(c)}{=}\frac{1-\beta}{1+\beta}{\rm Tr} \left[\left(\mU_{\F}^H {\rm diag}(\bp)\mathbf{C}_v^{-1}\mU_{\F}\right)^{-1}\right]\nonumber
\end{align}
where in (a) we used $\mU_{\F}^H\mU_{\F}=\mI$; in (b) we exploited Assumption 2, the uncorrelatedeness
 of the observation noise, the relation $\mathbb{E}\{\mD_{\S[l]}\mathbf{C}_v^{-1}\mD_{\S[l]}\}={\rm diag}(\bp)\mathbf{C}_v^{-1}$, and $\lim_{n\rightarrow\infty} \sum_{l = 1}^n\beta^{2(n-l)}=1/(1-\beta^2)$; finally, in (c) we have used (\ref{approx}).
This concludes the proof.

\section*{Acknowledgements}
We thank the authors of \cite{romero2016kernel} for having shared the code to reproduce the results obtained by their algorithm.

\balance
\bibliographystyle{MyIEEE}
\bibliography{refs}

\end{document}